\documentclass[journal]{IEEEtran}
\usepackage{graphicx}
\usepackage{multirow}
\usepackage{caption}
\usepackage{supertabular}
\usepackage{lineno,hyperref}
\modulolinenumbers[5]
\usepackage{makecell}
\usepackage[square, comma, sort&compress, numbers]{natbib}
\usepackage[labelformat=simple]{subcaption}

\usepackage{color}
\usepackage{amsfonts}
\usepackage{algorithm}
\usepackage{algpseudocode}
\usepackage{latexsym,bm}
\usepackage{amsmath,amssymb}
\usepackage{colortbl}
\usepackage{algorithmicx}
\usepackage{mathtools}
\ifCLASSINFOpdf
\else
\fi

\begin{document}
%
\title{Artificial intelligence-enabled detection and assessment of Parkinson's disease using multimodal data: A survey}

\author{
Aite Zhao, Yongcan Liu, Xinglin Yu, and Xinyue Xing
\thanks{This research was supported in part by National Natural Science Foundation of China under Grant No.62106117, China Postdoctoral Science Foundation under Grant No.2022M711741, and Natural Science Foundation of Shandong Province under Grant No.ZR2021QF084. }
\thanks{(Corresponding authors: Aite Zhao)}
}
\markboth{}%
{Shell \MakeLowercase{\textit{et al.}}: Bare Demo of IEEEtran.cls for IEEE Journals}

\maketitle

\begin{abstract}
The rapid emergence of highly adaptable and reusable artificial intelligence (AI) models is set to revolutionize the medical field, particularly in the diagnosis and management of Parkinson's disease (PD). Currently, there are no effective biomarkers for diagnosing PD, assessing its severity, or tracking its progression. Numerous AI algorithms are now being used for PD diagnosis and treatment, capable of performing various classification tasks based on multimodal and heterogeneous disease symptom data, such as gait, hand movements, and speech patterns of PD patients. They provide expressive feedback, including predicting the potential likelihood of PD, assessing the severity of individual or multiple symptoms, aiding in early detection, and evaluating rehabilitation and treatment effectiveness, thereby demonstrating advanced medical diagnostic capabilities. Therefore, this work provides a surveyed compilation of recent works regarding PD detection and assessment through biometric symptom recognition with a focus on machine learning and deep learning approaches, emphasizing their benefits, and exposing their weaknesses, and their impact in opening up newer research avenues. Additionally, it also presents categorized and characterized descriptions of the datasets, approaches, and architectures employed to tackle associated constraints. Furthermore, the paper explores the potential opportunities and challenges presented by data-driven AI technologies in the diagnosis of PD.


\end{abstract}

\IEEEpeerreviewmaketitle

\section{Introduction}
\IEEEPARstart{P}{arkinson's} disease (PD) is the fastest-growing neurological disorder worldwide. As of 2020, over 1 million people in the United States are affected by PD \cite{YangArtificial2022}, and China has approximately 3.5 million patients, a number expected to rise to 5 million by 2030 \cite{Utilization2023}. To date, no medication has been capable of reversing or preventing the disease's progression \cite{YangArtificial2022}. A critical challenge in PD's diagnosis, treatment, and management is the absence of an objective and efficient assessment tool \cite{10.1007/978-3-030-59716-0_61}. PD is typically diagnosed through clinical symptoms, predominantly hand tremors, speech and gait disorders, and other motor impairments. However, diagnosis relies heavily on the subjective evaluations of physicians, and the symptoms fluctuate dynamically before and after treatment, necessitating time-consuming visual observations and questionnaires. Early-stage patients may exhibit subtle limb changes that are hard to detect, impeding early medical intervention and risking delayed diagnosis, which could exacerbate the condition. Consequently, there is an urgent demand for a novel paradigm in medical artificial intelligence for diagnosing and assessing PD, with the goal of streamlining the conventional processes of PD diagnosis, treatment, and rehabilitation.

\begin{figure}[htp]
  \centering
  \includegraphics[width=8.5cm]{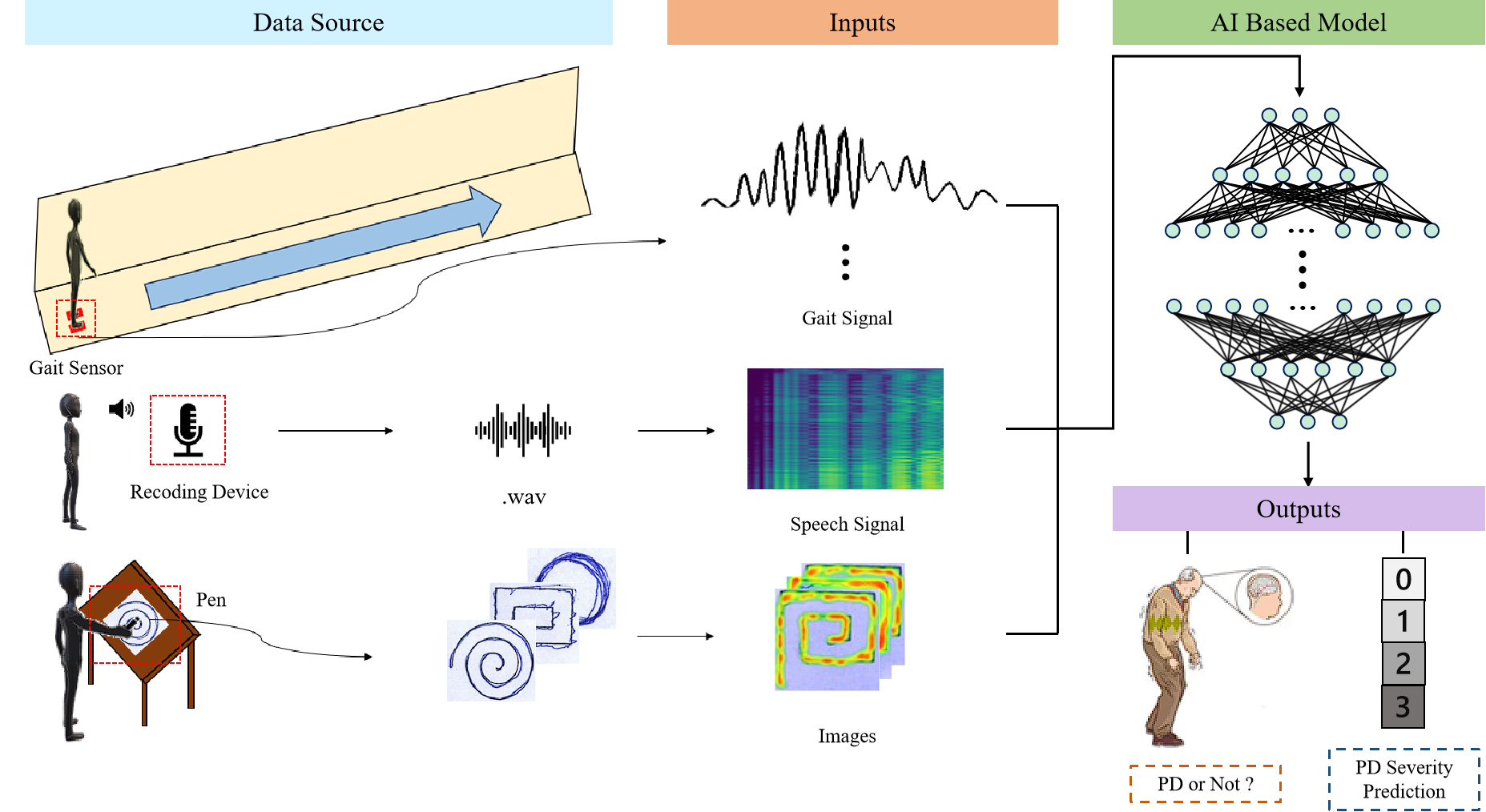}
  \caption{The process of PD multi-modality and multi-symptom detection and evaluation.}
  \label{fig_aisys}
\end{figure}

Effective strategies for tracking the progression of PD over time remain elusive. Current PD symptom assessments are heavily dependent on patient self-reports or the qualitative evaluations provided by clinicians. The diagnosis often involves the use of questionnaires such as the Movement Disorder Society's Unified Parkinson’s Disease Rating Scale (MDS-UPDRS) or the Hoehn-Yahr scale, both of which are semi-subjective and lack the sensitivity required to detect minor fluctuations in a patient's condition. Consequently, clinical trials for PD must span several years to ensure that changes in MDS-UPDRS scores are reported with adequate statistical confidence, a factor that escalates costs and prolongs the research timeline \cite{YangArtificial2022}. Thus, there is a pressing demand for an automated and objective method to assess disease progression, which could streamline clinical trials and enhance the timeliness and accuracy of PD management.

\begin{figure*}[ht]
  \centering
  \includegraphics[width=13cm]{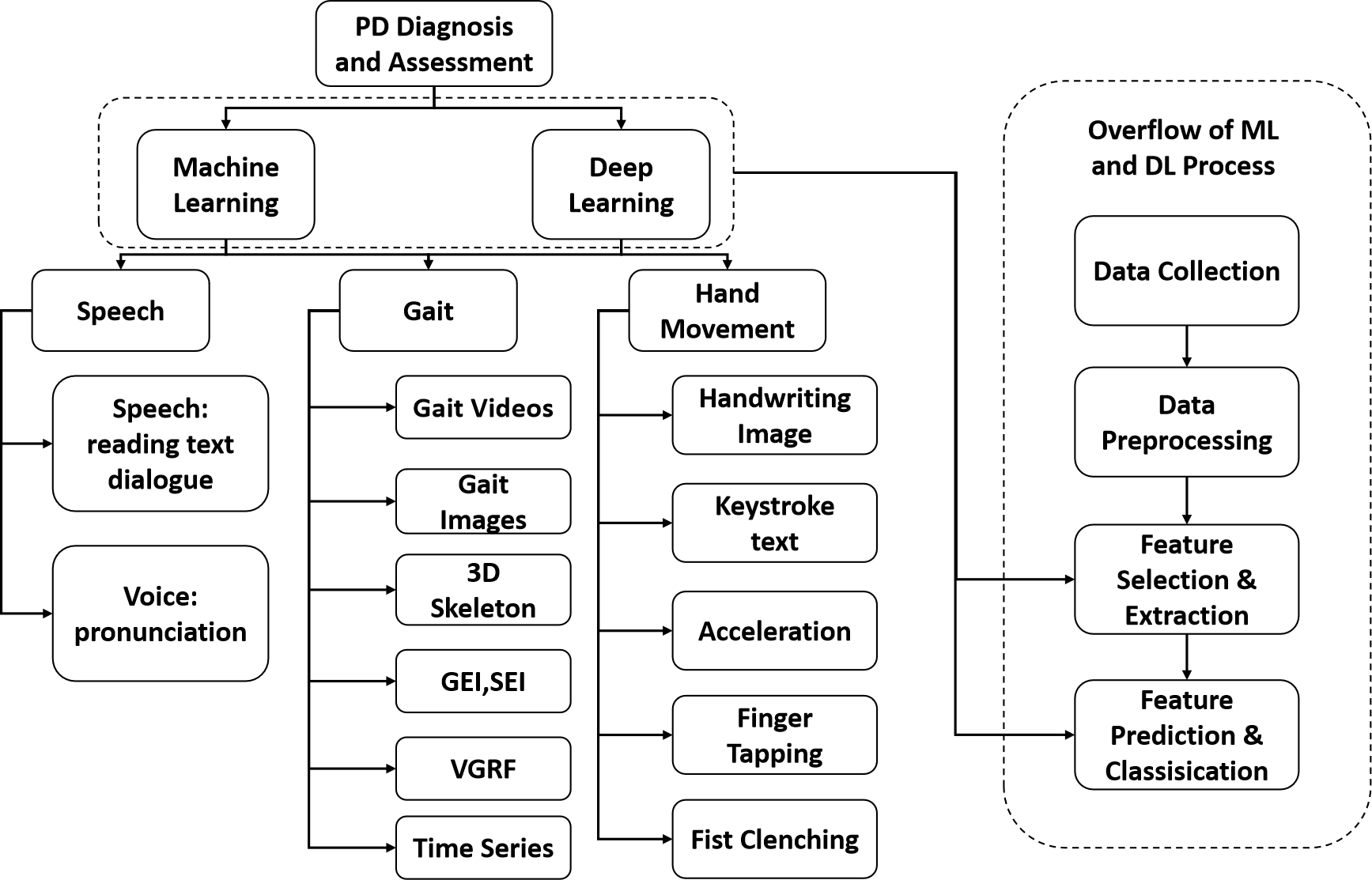}
  \caption{Classification of modalities for the diagnosis and assessment of Parkinson’s Disease (PD).}
  \label{fig_frame}
\end{figure*}
The existing literature has delved into various AI-assisted methodologies for the detection and assessment of PD, yet these have predominantly concentrated on singular modalities or discrete tasks within the diagnostic and therapeutic continuum of PD. This encompasses areas such as PD screening, the grading of symptom severity, and the tracking of specific biomarkers. For instance, the analysis and diagnosis based on PD brain imaging, the gauging of PD severity through gait force sensor data, the comprehension of PD-affected speech, and the evaluation of handwriting in PD patients have been explored. However, these studies are constrained by their reliance on a narrow range of data types and typically address only a single stage of PD's progression.

In this article, we present a detailed review of various machine learning (ML) and deep learning (DL)-based AI techniques applied to PD diagnosis and assessment. Our review synthesizes findings from a broad spectrum of published research, shedding light on emerging trends in PD diagnostics. As illustrated in Fig. \ref{fig_aisys}, the hybrid AI model is trained on various PD data through ML and DL techniques. Next, the model needs to access various sources of medical knowledge to perform medical reasoning tasks, thereby releasing rich features that can be used for downstream applications, such as severity grading for patient speech, gait and hand movement data. The subsequent sub-models then perform tasks or feedback that users can specify in real time. The article will proceed with a comprehensive overview of the integration of multi-source symptom data, highlighting the nuances and potentials of this approach in advancing PD diagnosis and assessment.

Gait abnormalities are a defining feature of motor symptoms in neurodegenerative diseases, with PD being no exception. PD is particularly characterized by distinctive changes in walking posture throughout the disease's progression, such as the well-known shuffling and freezing gaits. By employing machine learning techniques to analyze and decode the intricate dynamics and patterns within these abnormal gaits, we can achieve a more accurate differentiation among PD patients, taking into account the unique stages and varying levels of severity they present. The gait data of PD patients show considerable diversity, with each type of sensor capturing specific traits of movement patterns. This rich data can be harnessed to refine the diagnostic process, potentially leading to earlier and more precise identification of PD, as well as more personalized treatment strategies.
To collect gait data, employing visual sensors like cameras offers a straightforward approach to capturing the comprehensive walking patterns of individuals. Subsequently, these data are meticulously analyzed and processed through sophisticated automated learning algorithms.
Skeletal features serve as a pivotal reference in gait recognition, as they effectively filter out extraneous details, focusing solely on the joint coordinates of the human skeletal structure. This refined approach allows for a more precise identification of gait variations.
Furthermore, data gleaned from force sensors, exemplified by the vertical ground reaction force (VGRF), offer a nuanced perspective on the subtleties of lower limb dynamics. This approach has garnered considerable research attention across various regions globally, focusing on the analysis of gait through force sensor data.

PD often affects the hands of patients, frequently resulting in a characteristically flexed posture due to muscle rigidity, which impedes the full extension of the palms. Moreover, the small joints between the fingers can become hyperextended, further complicating the action of making a fist. To tackle these pathological manifestations, our project will systematically collect data on patients' hand movements. We will analyze the early impacts of PD on the kinematic characteristics of hand and finger movements and evaluate the severity of these impairments. As widely recognized, biomarkers indicative of PD can be identified through various modes of human-computer interaction, including the assessment of grip strength, the finger tapping test (FTT), and the observation of hand and finger movements.

Speech disorders in PD present as a distinctive complication, characterized by a sudden emergence and a gradual progression of communication barriers. Patients often encounter a reduction in vocal clarity and volume, hoarseness, and in severe cases, even a complete loss of voice. Some individuals may also develop stuttering, which severely hampers their social interactions. Moreover, uncontrollable vocal tremors and indistinct articulation can render patients incomprehensible to their colleagues and family, potentially leading to emotional distress and social withdrawal. The co-occurrence of speech disorders in Parkinson's patients exacerbates their suffering, significantly diminishing their quality of life and leaving their families feeling helpless and at a loss for support. Therefore, scientific research should be conducted on language disorders in different stages of PD, using speech data from PD patients to assist doctors in early diagnosis of the disease and assess its severity.

The primary objective of this study is to provide an in-depth analysis of the diverse methodologies employing ML and DL for the diagnosis and evaluation of PD. As illustrated in Fig. \ref{fig_frame}, the classification of modalities for diagnosing and assessing PD utilizes two artificial intelligence algorithms that operate on multimodal data derived from three distinct symptoms. Following the process of data analysis and processing, feature selection and extraction are conducted to yield classification and prediction outcomes that aid in the diagnosis and treatment of PD. By meticulously examining the extensive body of published research from a multitude of scholars, this study distills deeper insights into current research trends and underscores the evolving paradigms in PD diagnosis.

This work introduces various research results and studies on the use of machine learning and deep learning techniques for diagnosing PD published in various electronic database search engines (such as IEEE, Google Scholar, Web of Science, etc.) between 2014 and 2024. In recent years, it has been observed that the number of PD diagnostic articles based on deep learning techniques has steadily increased. From the literature review, it can be noted that most authors only investigated the contribution of AI methods to the diagnosis of PD. But this review compares and analyzes the experimental results of various researchers, provides algorithms for multimodal learning and analysis of multimodal datasets, and compares feature selection models, classifiers, and sensors, providing new perspectives for future work. Hence, the significance of this comprehensive review paper are as follows:

i. This article summarizes the contributions of traditional machine learning techniques and deep learning models to the analysis and processing of multimodal, multi symptom, and heterogeneous data.

ii. Emphasize the functionality and effectiveness of feature selection, extraction, and recognition models, as they play a crucial role in improving the accuracy of PD diagnostic assessment

iii. This study introduces a multimodal dataset of speech, gait, and hand movements. Enable readers to have a more detailed understanding of the acquisition process and experimental conditions of multimodal data.

vi.For researchers interested in developing PD prediction models through various AI based approaches in the future, this review paper may be a valuable source.

The structure of this article unfolds as follows: Initially, it presents an overview of the significance of PD, highlighting both its motor and non-motor symptoms. The subsequent section delves into the utilization of machine learning methodologies for the diagnosis and assessment of PD symptoms. Moving forward, the third segment offers an in-depth examination of various deep learning approaches employed in the diagnosis and evaluation of the condition. The fourth section introduces and scrutinizes publicly accessible datasets pertaining to the clinical manifestations of PD. Conclusively, the final section conducts a scholarly analysis of the principal challenges and future directions within this domain.

\section{PD Diagnosis and Assessment through ML-based Approaches}
This section evaluates the efficacy of machine learning techniques in analyzing three key clinical manifestations of Parkinson's disease, examining both singlemodal and multimodal analytical perspectives.
\begin{figure}[ht]
  \centering
  \includegraphics[width=8.5cm]{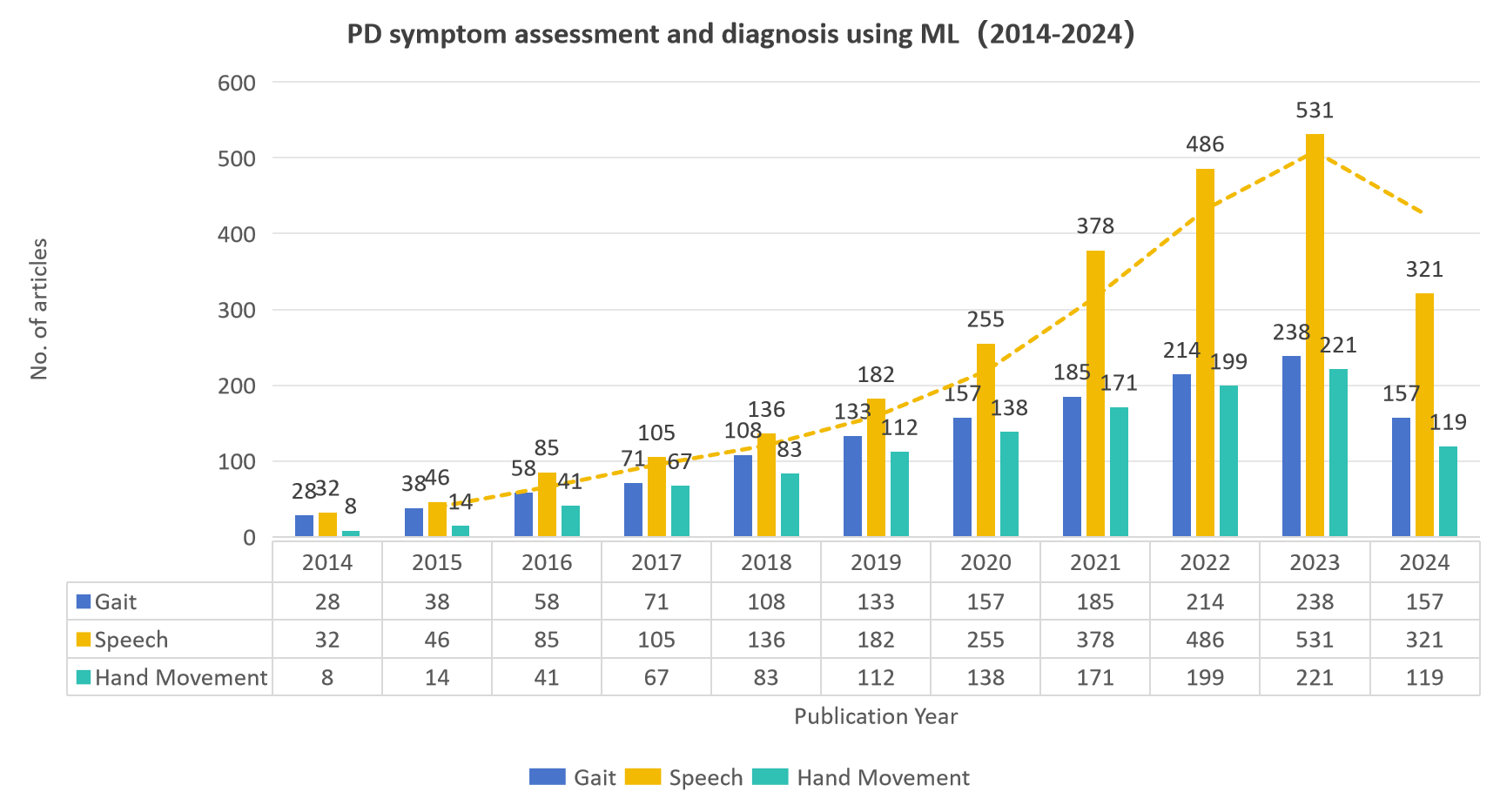}
  \caption{Single modal PD symptom assessment and diagnosis methods using ML (2014-2024).}
  \label{fig_mlpaper}
\end{figure}

By comparing the three biometric features and their symptom manifestations, we can see that from 2014 to 2024, machine learning algorithms mainly focused on speech recognition, and the number of new papers added each year has consistently ranked first over the past decade. This is also related to the singularity and ease of collecting voice data. There are relatively few papers on hand movement disorders and tremor detection, as the precision and subtle changes in hand or finger movements are difficult to grasp and closely related. Gait involves the entire human body and has various sources and modes, which determines its important position in the diagnosis and treatment of Parkinson's disease.

Machine learning is a branch of artificial intelligence that enables computer systems to learn from data and make decisions or predictions without the need for explicit programming. Machine learning techniques include supervised learning, unsupervised learning, semi supervised learning, and reinforcement learning. Each method has its specific application scenarios and advantages.

Machine learning can analyze the motor and non-motor symptoms of Parkinson's disease patients, such as tremors, rigidity, and bradykinesia; By analyzing patients' physiological and behavioral data, machine learning models can help doctors make more accurate diagnoses, track disease progression, and provide personalized treatment plans for patients. The application of machine learning in Parkinson's disease detection is becoming increasingly widespread. By analyzing large amounts of medical data, it improves the accuracy of diagnosis and provides better treatment and management plans for patients. With the advancement of technology, the application of machine learning in this field will become more profound and extensive in the future.

\subsection{Single Modal PD Symptom Analysis} 
Over the past decade, machine learning has emerged as a beacon of innovation in the realm of Parkinson's disease diagnosis and assessment. Especially in the processing of singlemodal normalized data, its characteristics of few parameters, high efficiency, and high accuracy make it dominant in PD data processing.

In recent years, the spotlight has been on the Support Vector Machine (SVM) as a prominent machine learning method. SVM is a versatile linear classifier that excels in binary classification tasks through the application of supervised learning. Its emphasis on precision in classification makes it a pivotal tool in the diagnosis and evaluation of Parkinson's disease, addressing the complexities of this condition with remarkable efficacy.

\begin{table*}[!htp]
\centering 
\caption{Summarises the diagnosis and assessment of PD based on speech data using machine learning techniques.}
\scalebox{0.8}{
\begin{tabular}{l|l|l|l|p{8cm}|l|l|l}
  \hline
  \hline
 References&Dataset&No. of subjects&Feature extraction/selection&Classifier&Accuracy&Sensitivity&Specificity\\
\hline
\cite{shahbakhi2014speech}&Speech&23-PD, 8-HC&Fundamental, frequency or pitch&SVM&94.50\%&*&*\\
\cite{benba2015voiceprints}&Voice&17-PD, 17-HC&MFCC&SVM&91.17\%&*&*\\
\cite{benba2015detecting}&Voice&17-PD, 17-HC&MFCC&SVM(RBF)&91.18\%&*&*\\
\cite{soumaya2019diagnosis}&Speech&20-PD, 18-HC&MFCC(DWT)&Linear SVM&72\%&80\%&65\%\\
\cite{solana2020automatic}&Voice&188-PD, 64-HC&TQWT and MFCC&KNN,SVM,RF,MLP&94.7\%&98.4\%&92.68\%\\
\cite{soumaya2021detection}&Speech&20-PD, 14-HC&DWT, MFCC&SVM&91.18\%&*&*\\
\cite{rahman2021parkinson}&Speech&60-PD, 100-HC&MFCC&LDA-SVM&88\%&73.33\%&84\%\\
\cite{9316078}&Voice&23-PD, 8-HC&Frequencey, Jitter and Shimmer&NB,LR,KNN,and RF&90.2\%&*&*\\
\cite{hawi2022automatic}&Voice&188-PD, 64-HC&MFCC and Long-term feature&RF&88.84\%&*&*\\
\cite{indu2023modified}&Voice&20-PD, 20-HC&Frequencey, Jitter and Shimmer&KNN&94.5\%&*&*\\
\cite{pramanik2021machine}&Voice&188-PD, 64-HC&MFCC, Wavelet and TQWT&Decision forests&94.12\%&94\%&94\%\\
\cite{ouhmida2022parkinson}&Voice&40-PD, 40-HC&Acoustic feature&SVM, LR, Discriminant Analysis, KNN, DT, RF, Bagging tree, NB, and AdaBoost&97.22\%&94.44\%&100\%\\
\cite{zhang2016classification}&Speech&20-PD, 20-HC&Acoustic feature&ensemble learning&88\%&97\%&88\%\\
\cite{liu2021local}&Speech&20-PD, 20-HC&Manual feature&ELM+SVM&97.50\%&*&*\\
\cite{barukab2022analysis}&Speech&188-PD, 64-HC&753 features&AdaBoost, RF, and DT&95.20\%&98.30\%&62.20\%\\
\cite{oung2018evaluation}&Speech&20-PD, 10-HC&MFCC,LPCC,LPC,WLPCC&KNN,PNN&92.63\%&*&*\\
 \hline
 \hline
\end{tabular}}
  \label{tb_ml_speech}
\end{table*}

Firstly, SVM has made significant contributions to speech recognition and classification for PD patients. Shahbakhi \textit{et al.} proposed a new algorithm for Parkinson's disease diagnosis based on biomedical speech signal analysis, using genetic algorithm to select optimized features from all extracted features. Then, SVM was used to distinguish between healthy individuals and Parkinson's disease patients \cite{shahbakhi2014speech,benba2015voiceprints,benba2015detecting,soumaya2019diagnosis}. The extraction of MFCC features from general speech data was a key feature widely used for extracting audio signals, as it can effectively represent the spectral characteristics of audio signals, especially performing well in noisy environments. SVM was also used to distinguish samples represented by MFCC features \cite{kuresan2019fusion,solana2020automatic,soumaya2021detection,rahman2021parkinson,boualoulou2023comparison,hawi2022automatic}. It appeared in the decision layer for feature classification, and other classifiers with the same functionality included Logistic Regression (LR), Naive Bayes (NB), k-nearest neighbors (KNN), multi-layer perceptron (MLP), Decision Tree (DT), Random Forest (RF) and XGBoost, which can process concatenated MFCC features to provide accurate classification results \cite{9316078,rahman2021parkinson,hawi2022automatic, indu2023modified,pramanik2021machine,ouhmida2022parkinson}. The ensemble learning algorithm \cite{zhang2016classification,nissar2019voice,liu2021local,9793048,barukab2022analysis,shastry2023ensemble} that collaborated with these classifiers using voting decisions to distinguish speech features had greater advantages. For example, an ensemble learning algorithm consisting of AdaBoost, random forest, and decision tree \cite{barukab2022analysis} developed on an imbalanced speech dataset selected 10 best features and obtained 90\% of the results for the optimal ensemble classifier and comprehensive evaluation. In addition to utilizing MFCCs, a variety of other acoustic features, such as Tunable Q-factor Wavelet Transform (TQWT), Linear Predictive Cepstral Coefficients (LPCC) were concurrently employed as inputs for classifiers, which were designed to discern the progression and variations in vocal impairments across various types of Parkinson's Disease (PD) patients \cite{celik2023proposing,zlotnik2015random,oung2018evaluation}. 

As demonstrated in Table \ref{tb_ml_speech}, 16 literature on PD speech recognition were investigated, including raw data on voice and speech. The maximum number of participants is 188 PD patients and 64 healthy subjects, collected under different experimental conditions. Most of these data were preprocessed to extract features such as Mel-frequency cepstral coefficients (MFCC), Discrete Wavelet Transform (DWT), Tunable Q-Factor Wavelet Transform (TQWT), Linear Prediction cepstral coefficients (LPCC), Linear Predictive Coding (LPC), weighted linear prediction cepstral coefficients (WLPCC), etc., and then fed into support vector machine (SVM), random forest (RF), naive Bayes (NB), k-nearest neighbour (KNN), multi-layer perceptron (MLP), Adaboost, logistic regression (LR), decision tree (DT), extreme learning machine (ELM), and ensemble learning for classification. The majority of the predicted results obtained exceeded 90\%. MFCC is a popular feature for speech processing due to its human-like sensitivity to frequency, key information extraction like pitch and timbre, and robustness to speaker volume and noise. It also reduces data dimensions while preserving essential details, enhancing algorithm efficiency and reducing computational load. These traits make MFCC a preferred choice for tasks like speech recognition and synthesis.

\begin{table*}[!htp]
\centering 
\caption{Summarises the diagnosis and assessment of PD based on  gait  data using machine learning techniques.}
\scalebox{0.8}{
\begin{tabular}{l|l|l|p{5cm}|p{5cm}|l|l|l}
  \hline
  \hline
 References&Dataset&No. of subjects&Feature extraction/selection&Classifier&Accuracy&Sensitivity&Specificity\\
\hline
\cite{kim2021abnormal}&Gait&*&Integrated gait features&KNN,\textbf{SVM}&96.52\%&*&*\\
\cite{zhao2018hybrid}&Gait&93-PD, 73-HC&VGRF,CNN,LSTM,temporal features&\textbf{LSTM,CNN}&98.88\%&*&*\\
\cite{khan2021novel}&Gait&19-PD&Silhouette, and temporal features&\textbf{SVM}&70.83\%&* &*\\
\cite{wahid2015classification}&Gait&23-PD, 26-HC&Spatial-temporal gait features&KFD, BA, KNN, SVM, \textbf{RF}&92.6\%&96\%&100\%\\
\cite{shetty2016svm}&Gait&15-PD,16-HC, 20-HD,13-ALS&Stride,swing and stance&\textbf{SVM(RBF)}&83.33\%&85.71\%&75\%\\
\cite{abdulhay2018gait}&Gait&93-PD, 73-HC&Gait,tremor,kinetic,and temporal features&\textbf{SVM}&92.7\%&*&*\\
\cite{alkhatib2020machine}&Gait&29-PD, 18-HC&VGRF,COP,load distribution&\textbf{LDA},QDA&95\%&*&*\\
\cite{balaji2020supervised}&Gait&93-PD, 73-HC&Temporal and spatial features&DT, SVM, EC&*&*&*\\
\cite{chavez2022vision}&Gait&*&Stride,swing and double support&SVM, KNN,\textbf {GB}&99\% &97\%&99\%\\
\cite{chen2022computer}&Gait&7-HC&Gait cycle,joint angles&\textbf{SVM},KNN, LSTM,CNN&94.9\%&*&*\\
\cite{loureiro2020using}&Gait&10 mix&GEI, SEI, joint angles,CNN features&\textbf{VGG-19 CNN}, LDA, SVM&98.5\%&*&*\\
\cite{munoz2022machine}&Gait&30-PD, 30-HC&Gait spatiotemporal features, swing magnitude, asymmetry&\textbf{RF}, SVM, DT, NB, LR&84.5\%&*&*\\
\cite{orphanidou2018predicting}&Gait&93-PD, 73-HC&Time domain, non-linear and frequency domain features&MLP,RF,XGB,SVML,
KNN,NB,\textbf{SVMP}&91\%&89\%&95\%\\
\cite{gong2020novel}&Gait&30 mix&Gait energy images&\textbf{OSVM}&97.33\%&98.25\%&95.89\%\\
\cite{borzi2023context}&Gait&81-PD&Temporal and spectral features&LR, RF, CNN&* &96\%&93\%\\
\cite{kleanthous2020new}&Gait&10-PD&Time and frequency domain features&RF, NN, \textbf{SVM(RBF)},  XGBoost&79.85\% &*&*\\
 \hline
 \hline
\end{tabular}}
  \label{tb_ml_gait}
\end{table*}

Furthermore, these methods had not only performed well in the field of PD speech recognition, but they also exhibited stable performance in the realm of PD gait recognition. Although PD gait data were multi-sensor and multimodal, mostly from conditional laboratory acquisition, a lot of work had been focused on a single modality. Gait data generally included three-dimensional joint points of skeleton, ground reflection force, and image data \cite{kim2021abnormal,zhao2018hybrid,khan2021novel}. Most machine learning classifiers had determined the severity of PD or detected PD in a population by learning gait patterns or data characteristics \cite{wahid2015classification,shetty2016svm,abdulhay2018gait,alkhatib2020machine,balaji2020supervised}. An automatic gait classification system \cite{balaji2020supervised} had been proposed for identifying the severity level of PD, using four supervised machine learning classifier algorithms, namely, DT, SVM, ensemble classifier (EC), and Bayes classifier (BC), obtaining an accuracy of 99.4\% on the GRF gait dataset. Machine learning strategies had also been employed to classify PD gait and healthy controls \cite{wahid2015classification} based on GRF data, using RF, SVM, KNN, and kernel Fisher discriminant. ML algorithms had also been very sensitive to gait skeleton data \cite{kim2021abnormal,chavez2022vision,chen2022computer}. Based on 3D-depth skeleton data, Kim \textit{et al.} \cite{kim2021abnormal} had built a KNN classifier and an SVM classifier to classify abnormal gait from a walking person. Chavez \textit{et al.} \cite{chavez2022vision} had presented the utilization of KNN, SVM, and gradient boosting (GB) classifiers in distinguishing well-established gait features from the skeleton. For the RGB gait images and videos, machine learning approaches could capture changes of each frame and directly classify the PD feature \cite{loureiro2020using,khan2021novel,munoz2022machine}. The SVM had classified the levels with a promising area under the ROC of 80.88\% on a video dataset of PD gait \cite{khan2021novel}. DT, NB, and RF had worked together on a PD RGB-D camera dataset collected using Kinect \cite{munoz2022machine}. There had also been some machine learning methods applied to detect freezing of gait (FOG) and analyze gait energy images  \cite{orphanidou2018predicting,gong2020novel,borzi2023context,kleanthous2020new}.

Table \ref{tb_ml_gait} recaps the diagnosis and assessment of PD using machine learning techniques to analyze gait data. The table highlights in bold the algorithms that achieved the highest accuracy rate among all the studies listed. It reflects the critical role of feature extraction and classifier selection in improving diagnostic performance. For example, studies using spatial-temporal features are able to capture subtle variations in gait, and deep learning classifiers such as CNN and LSTM can process these complex features to achieve high diagnostic accuracy. In the studies referenced as \cite{abdulhay2018gait,balaji2020supervised,orphanidou2018predicting}, the use of extensive datasets has allowed for the application of diverse algorithms, including SVM, KNN, DT, RF, and ensemble methods like XGBoost and MLP. The performance of these algorithms on larger datasets is especially noteworthy, as it indicates their potential for real-world applicability and reliability in diagnosing PD.

\begin{table*}[!htp]
\centering 
\caption{Summarises the diagnosis and assessment of PD based on  hand movement data using machine learning techniques.}
\scalebox{0.8}{
\begin{tabular}{l|p{2cm}|l|p{6cm}|p{4cm}|l|l|l}
  \hline
  \hline
 References&Dataset&No. of subjects&Feature extraction/selection&Classifier&Accuracy&Sensitivity&Specificity\\
\hline
\cite{islam2024review}&Voice,handwriting&*&Voice dataset, handwriting dataset features&SVM,RF,KNN,\textbf{CNN}&95\%&*&*\\
\cite{impedovo2018dynamic}&Handwriting&37-PD,38-HC&Kinematic,Spatio-Temporal features &KNN,SVM(RBF),NB, LDA,\textbf{RF},ADA&73.38\%&68.87\%&88.89\%\\
\cite{pereira2015step}&Handwriting&37-PD,18-HC&Visual-Based features, MRT, statistical features&\textbf{NB},PF,SVM(RBF)&78.9\%&*&*\\
\cite{ranjan2023detection}&Handwriting&*&HOG&\textbf{RF}&86.67\%&*&*\\
\cite{kotsavasiloglou2017machine}&Handwriting&24-PD,20-HC&NVV,SDV,MV&*&91\%&88\%&95\%\\
\cite{castrillon2019characterization}&Handwriting&55-PD,94-HC&Neuromotor, kinematic,nonlinear dynamic&KNN,MLP,\textbf{SVM}&97\%&*&*\\
\cite{guarin2024characterizing}&Handwriting&66-PD,24-HC&Movement Amplitude,Speed,Amplitude Decay&Multiclass, Ordinal Binary and \textbf{Tiered Binary Classification}&86\%&*&*\\
\cite{yang2022automatic}&Finger tapping and postural stability &579-PD&Tapping rate, times,  amplitude variation,  body straightness&\textbf{DNN}&88\%&*&*\\
\cite{yu2023clinically}&Finger tapping&50-PD&Demographic factors and kinematic features &\textbf{DT}&80\%&*&*\\
 \hline
 \hline
\end{tabular}}
  \label{tb_ml_hand}
\end{table*}

According to the UPDRS, hand motor symptoms generally occurred in the early stages of Parkinson's disease, characterized by static tremors and slow, clumsy movements. The severity of these symptoms was determined by the speed, amplitude, and rhythm of finger tapping, hand opening and closing, and hand spinning movements, or changes before and after rehabilitation training. Hand tremors directly affected the stability of handwriting \cite{islam2024review}. Machine learning algorithms learned the movement trajectory and handwriting of hand movements to distinguish between PD patients of different severity levels or to conduct early screening for potential PD \cite{impedovo2018dynamic,pereira2015step,ranjan2023detection,kotsavasiloglou2017machine,castrillon2019characterization}. A random forest classifier \cite{ranjan2023detection} was employed for the detection of Parkinson's disease among patients, using two types of handwriting analysis, i.e., spiral and wave drawings. The RF, NB, boosted tree, logistic regression, and SVM classified Normalized Velocity Variability (NVV), the velocity’s Standard Deviation (SDV), and Mean (MV), and the signal Entropy (ETP), of the PD line-drawing; NB had been verified to achieve the highest accuracy \cite{kotsavasiloglou2017machine}. In addition to handwriting, finger-tapping and fist-clenching exercises were also important references for evaluating PD during the rehabilitation stage \cite{guarin2024characterizing,yang2022automatic,khan2014computer,yu2023clinically,he2023effective,serrano2024estimation,TODOROV2024146}. For example, SVM was used on PD video footages for UPDRS motor examination of finger-tapping (UPDRS-FT), which could evaluate the severity of PD \cite{khan2014computer}. XGBoost was proposed to distinguish Parkinsonian rest tremor and fist-clenching by analyzing brain activity, providing better results than other linear models \cite{TODOROV2024146}.

Table \ref{tb_ml_hand} summarises most works for the diagnosis and assessment of PD based on hand movement data using machine learning techniques. The table highlights in bold the highest reported accuracy achieved within the listed studies. A distinctive aspect of the table is the variety in the combination of feature extraction methods and classifiers. For example, the study \cite{impedovo2018dynamic} uses kinematic and spatio-temporal features with a range of classifiers, including KNN, SVM with RBF kernel, NB, LDA, RF, and ADA, achieving an accuracy of 73.38\%. This demonstrates the potential of combining specific features with multiple classifiers to enhance diagnostic accuracy. The study \cite{yang2022automatic} involved 579 PD patients and used features such as tapping rate, timing, amplitude variation, and body straightness with a DNN classifier, achieving an accuracy of 88\%.

\subsection{Multimodal PD Symptom Analysis}

Machine learning algorithms were not only suitable for single mode and single symptom but also suitable for multimodal data and multi-symptom analysis, as long as different modes were mapped in the same feature space \cite{nahiduzzaman2020machine,li2024image,bi2020multimodal}. Li \textit{et al.} introduced a multi-modal data image encoding and fusion approach for diagnosing depression in PD patients, using a multi-modal dataset encompassing motion, facial expression, and audio data \cite{li2024image}. Fisher feature classifiers and deep canonical correlation analysis (DCCA) were applied on a multimodal PD gait dataset, fusing VGRF signal and temporal data for PD detection \cite{9376702}. In the study of \cite{bi2020multimodal}, a correlation analysis method was used to detect the relationship between brain regions and genes, using RF and DT for classification on functional magnetic resonance imaging (fMRI) and single nucleotide polymorphisms (SNPs). Another study \cite{10546910} compared Random Forest, Decision Tree, Logistic Regression, Gradient Boosting, Support Vector Machine, Stacking, and Bagging Ensemble classifiers to two standard benchmark datasets of PD. However, these methods only considered a single symptom or task of Parkinson's disease and did not provide diagnosis and treatment plans based on multimodal data of multiple symptoms like a professional doctor \cite{makarious2022multi}. In this study, Huo \textit{et al.} \cite{huo2020heterogeneous} used machine learning algorithms to quantitatively evaluate bradykinesia, stiffness, and tremor in PD patients and tested their ability to predict the UPDRS score on a system embedded with a force-sensor, three inertial measurement units, and four custom mechanomyography sensors.

Although machine learning methods have emerged as essential tools for decision-making in various Parkinson's disease (PD) detection tasks, their limitations lie in analyzing and interpreting similar data, as well as distinguishing various salient features. The extraction and description of these salient features depend on the learning capabilities of deep neural networks from different perspectives.

\section{PD Diagnosis and Assessment through DL-based Approaches}

\begin{figure}[ht]
  \centering
  \includegraphics[width=8.5cm]{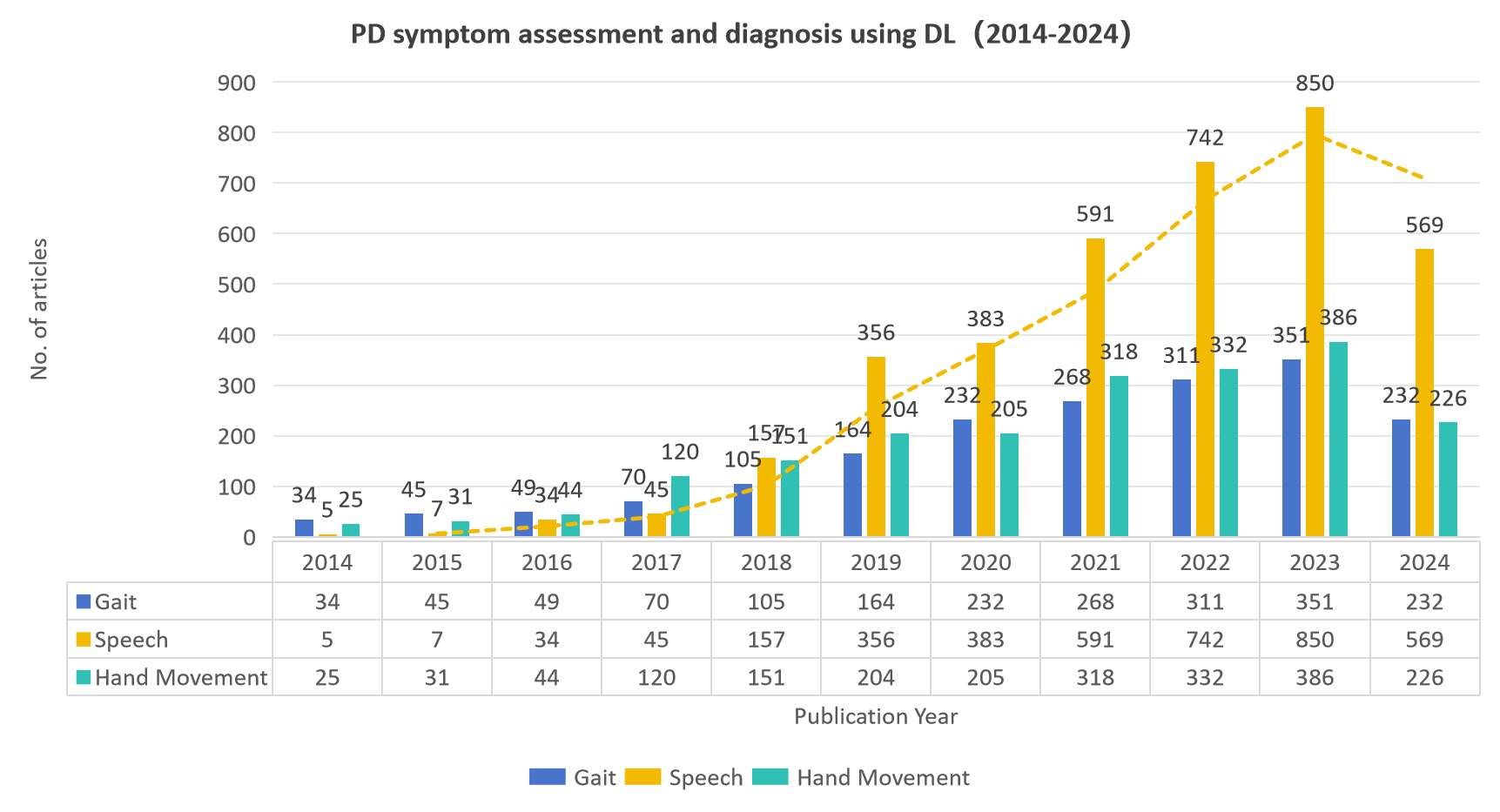}
  \caption{Single modal PD symptom assessment and diagnosis methods using DL (2014-2024).}
  \label{fig_dlpaper}
\end{figure}

As illustrated in Fig. \ref{fig_dlpaper}, deep learning technology rapidly developed between 2015 and 2020 and required sufficient training data. It is noteworthy that few deep learning models for speech recognition were developed in 2014 and 2015. LSTM entered a high-speed development stage from 2014 to 2015 and was applied in various time series fields. Speech data, being temporal in nature, are well-suited for LSTM analysis. The development of algorithms such as NLP led to a steady increase in the number of related papers, which reached a peak in 2023. The number of studies on the other two symptoms is also high, as most hand movement and gait data are collected using visual sensors and presented in the form of images and videos. Convolution-based deep learning algorithms can directly extract spatial features from two-dimensional and three-dimensional data, which contain more structural information and thus require more complex algorithms for processing. The development of temporal networks, convolutional networks, and attention mechanisms not only laid the foundation for models like Transformers but also fueled the growth of deep learning algorithms, leading to a higher average number of publications than machine learning algorithms in the 2020s.

\subsection{Single Modal Learning for PD Symptom} 

This section investigates the impact of gait, hand movements, and speech performance based on single-data (from multi-sensor sources) on the diagnosis and treatment of Parkinson's disease.

\begin{table*}[!htp]
\centering 
\caption{Summarises the diagnosis and assessment of PD based on speech data using Deep learning techniques.}
\scalebox{0.8}{
\begin{tabular}{l|l|l|p{6cm}|l|l|l|l}
\hline
\hline
References&Dataset&No. of subjects&Feature extracted model&Classifier&Accuracy&Sensitivity&Specificity\\
\hline
\cite{asdadasda}&Speech&*&15 phonological features&RNN, GRUs&76\%&*&*\\
\cite{Hernandez2022CrosslingualSS}&Speech&70-PD, 70-HC&XLSR-PD&Conformer encoder and Transformer decoder&87.1\%&*&*\\
\cite{bhati2019lstm}&Speech&52-PD, 56-HC&LSTM-based siamese network&Feed forward NN&96.2\%&98.1\%&98.1\%\\
\cite{gunduz2019deep}&Voice&188-PD, 64-HC&TQWT, MFCC and Concat&CNN&86.9\%&*&*\\
\cite{vasquez2017convolutional}&Speech&50-PD, 50-HC&STFT, CWT and CNN&CNN&89\%&*&*\\
\cite{zhang2018deepvoice}&Voice&500-PD, 500-HC&Joint Time-Frequency Analysis algorithm&CNN&90.45\%&84\%&95\%\\
\cite{zhao2024triplet}&Voice&28-PD, 20-HC&MFCC&TmmNet&99.91\%&99.94\%&*\\
\cite{hemmerling2023vision}&Voice&104-PD, 77-HC&Mel-spectrograms&Vision Transformer&75.96\%&73.53\%&78.05\%\\
\cite{nijhawan2023novel}&Voice&188-PD, 64-HC&753 vocal features&Vocal Tab Transformer&*&90.378\%&*\\
\cite{fang2020parkinsonian}&Speech&34-PD, 34-HC&128 MFCCs&CNN, LSTM, \textbf{E2E}&94.5\%&*&*\\
\cite{mehra2024deep}&Speech&55 subjects&Swin transformer&BiLSTM-GRU&97.64\%&*&*\\
\cite{jeong2024exploring}&Speech&100-PD, 100-HC&128-d log Mel filterbank features, EfficientNet&AST, \textbf{PSLA}&92.15\%&91.53\%&92.79\%\\
\cite{klempivr2023evaluating}&Speech&28-PD, 22-HC&Wav2vec&Ensemble random forest&95\%&97\%&*\\
\hline
\hline
\end{tabular}}
\label{tb_dl_speech}
\end{table*}

In recent years, a large number of deep models extracted MFCC features from speech disorder data, driving in-depth research on PD \cite{9850832,Hernandez2022CrosslingualSS}. Between 2014 and 2019, due to the popularity of CNNs (Convolutional Neural Networks) and LSTMs (Long Short-Term Memory networks) and their widespread application in other pattern recognition fields, PD speech, characterized by its time-domain and frequency-domain features, could be served as input for these models to extract temporal and spatial features \cite{bhati2019lstm,gunduz2019deep,vasquez2017convolutional}. These two were also combined into a fusion model to extract spatiotemporal features of PD speech simultaneously \cite{zhang2018deepvoice}. The LSTM variant, GRU \cite{asdadasda}, was also employed to assess speech impairments by computing static features from a complete utterance. After 2020, some complex models and Transformer-based learning frameworks gradually emerged, and the strategy of incorporating attention mechanisms made PD speech recognition more accurate \cite{van2024innovative,zhao2024triplet,hemmerling2023vision,nijhawan2023novel,fang2020parkinsonian,klempir2024analyzing1,chowdary2023few}. Some transformer-based hybrid models also emerged, which performed better than single models. For example, the use of BiLSTM-GRU fusion with Swin transformer achieved a 97.64\% classification effect on spoken language of patients with pronunciation disorders \cite{mehra2024deep}. The triple multimodal network \cite{zhao2024triplet} that simultaneously integrated LSTM, GRU, CNN, and transformers represented a leap forward in multi-model fusion for PD speech signal processing. It was applicable to multiple PD speech datasets and, for the first time, comprehensively considered the spatiotemporal features, salient regions, and time-frequency domain features of PD speech. Moreover, there were also some DL-based PD speech assessment systems that had been put into PD self-test and rehabilitation processes \cite{Hernandez2022CrosslingualSS,jeong2024exploring,jiang2023self}. Abner \textit{et al.} \cite{Hernandez2022CrosslingualSS} explored the usefulness of using Wav2Vec self-supervised speech representation as the speech feature of dysarthria in training ASR systems and used a transformer-based context network for feature representation and classification. Due to the limited amount of PD speech data, self-supervised and unsupervised models flourished \cite{jiang2023self,klempivr2023evaluating,klempir2024analyzing1,zhou2023risevi}. For instance, Mian \textit{et al.} proposed an unsupervised autoencoder for PD speech feature selection and passed the compressed features to supervised machine-learning (ML) algorithms. From then on, pre-trained models gradually became the basis for PD speech feature extraction and parameter fine-tuning.

Table \ref{tb_dl_speech} summarizes the diagnosis and assessment of PD based on speech data using deep learning techniques. It presents an overview of the dataset, the feature extraction model, the classifier model, and performance metrics—including accuracy, sensitivity, and specificity—for each referenced study. Bold text indicates the best-performing model in that study, and the performance measures are for that model. Based on the data presented in the table, it can be inferred that attention-based models (e.g., Transformer) and LSTM-based models generally outperform traditional models (e.g., CNN, RNN) across a range of speech datasets. Most achieved more than 92\% accuracy and more than 90\% sensitivity. For the diagnosis and evaluation of PD on speech datasets, researchers seem more inclined to apply deep learning technology in the classification stage, while for feature extraction, common features such as MFCC and TQWT are used. Some studies use the Transformer model for feature extraction and achieve better results, but this does not mean that common features are not suitable for speech data. In the table, the TmmNet model achieves 99.91\% accuracy and 99.94\% sensitivity by inputting MFCC features. Additionally, efforts in feature extraction using large deep learning models such as Wav2vec may be more beneficial for speech diagnosis of PD. Even if an ensemble random forest is used as the classifier, it still achieves a commendable performance of 95\% accuracy and 97\% sensitivity.

\begin{table*}[!htp]
\centering 
\caption{Summarises the diagnosis and assessment of PD based on gait data using Deep learning techniques.}
\scalebox{0.9}{
\begin{tabular}{l|l|l|l|l|l|l|l}
\hline
\hline
References&Dataset&No. of subjects&Feature extracted model&Classifier&Accuracy&Sensitivity&Specificity\\
\hline
\cite{ZHAO201891}&Gait&15-PD, 16-HC&Dual Channel LSTM&Softmax&97.33\%&*&*\\
\cite{TIAN2022116}&Gait&9 subjects&AGS-GCN&GAP and Softmax&100\%&*&*\\
\cite{ajay2018pervasive}&Gait&26-PD, 23-HC&Analysis of gait features from video&DT&93.75\%&*&*\\
\cite{el2020deep}&Gait&93-PD, 73-HC&1D-Convnets&FC&98.7\%&97.8\%&99.2\%\\
\cite{zou2020deep}&Gait&18 subjects&CNN and LSTM&FC&93.7\%&*&*\\
\cite{zhao2021multimodal}&Gait&15-PD, 16-HC&SFE, CorrMNN&Multi-Switch Discriminator&99.86\%&*&*\\
\cite{balaji2021automatic}&Gait&93-PD, 73-HC&LSTM&FC and Softmax&98.6\%&96.20\%&98.08\%\\
\cite{kumar2023parkinson}&Gait&93-PD, 73-HC&CNN, CNN-LSTM&\textbf{CNN}, CNN-LSTM&95\%&*&*\\
\cite{info14020119}&Gait&3996-PD, 12216-HC&Conv block and FuseLG block&FuseLGNet&99.78\%&*&*\\
\cite{9956330}&Gait&93-PD, 73-HC&Transformer&FC&95.2\%&98.1\%&86.8\%\\
\cite{CHERIET2023107344}&Gait&43 subjects&Transformer&Multi-Speed Transformer&96.9\%&96.9\%&97.1\%\\
\cite{sun2022transformer}&Gait&*&Transformer, GAU&Enhanced Transformer&97.4\%&*&*\\
\cite{naimi2023hct}&Gait&93-PD, 73-HC&1D-ConvNets, Transformer&ConvNet-Transformer&97\%&98.7\%&86.1\%\\
\cite{wu2022multi}&Gait&52 subjects&Transformer&Multi-Level Fine-Tuned Transformer&99.83\%&*&*\\
\cite{wu2023attention}&Gait&93-PD, 73-HC&Transformer&FC and Softmax&98.86\%&98.86\%&*\\
\hline
\hline
\end{tabular}}
\label{tb_dl_gait}
\end{table*}

Furthermore, a number of intelligent automatic learning approaches were widely adopted in gait data-based PD detection. The gait recognition algorithms based on GCN (Graph Convolutional Network), CNN (Convolutional Neural Network), and LSTM (Long Short-Term Memory) networks remained the mainstream approach, and researchers paid close attention to gait images, force sensing signals, and accelerations \cite{2022Sensor,ZHAO201891,TIAN2022116,ajay2018pervasive,el2020deep,zou2020deep,zhao2021multimodal,balaji2021automatic,kumar2023parkinson}. Among them, CNN and LSTM were the most common methods due to their effectiveness in extracting spatiotemporal features. Kumar \textit{et al.} \cite{kumar2023parkinson} used CNN to diagnose PD and a CNN-LSTM model using the Hoehn \& Yahr rating scale led to severity rating prediction. An LSTM classifier was presented for gait-based PD diagnosis and severity rating \cite{balaji2021automatic}. The emergence of the Transformer also promoted the development of gait recognition in PD detection \cite{info14020119,9956330,CHERIET2023107344,naimi20241d,chen2023fuselgnet,sun2022transformer,naimi2023hct,wu2022multi}. For instance, Zhao \textit{et al.} had been deeply involved in the field of severity assessment and detection of Parkinson's disease based on Transformer models for a long time and had made outstanding contributions \cite{wu2022multi,chen2023fuselgnet,wu2023attention}. A Multi-Level Fine-tuned Transformer \cite{wu2022multi} was trained on sequential gait data to capture and analyze the overall spatiotemporal information for the recognition of normal and abnormal gait. Moreover, they also designed a FuseLGNet \cite{chen2023fuselgnet} that fused local and global information of the gait data for PD detection. Considering the temporal feature of the gait cycle, an attention-based temporal network \cite{wu2023attention} was proposed for the time series of GRF (Ground Reaction Forces) for PD severity rating.

Table \ref{tb_dl_gait} summarizes the diagnosis and assessment of PD based on gait data using deep learning techniques. It presents an overview of the dataset, the feature extraction model, classifiers, and performance metrics—including accuracy, sensitivity, and specificity—for each referenced study. Bold text indicates the best-performing model in that study, and the performance measures are for that model. Deep learning models for Parkinson's disease (PD) diagnosis based on gait data have demonstrated satisfactory performance, with most models exceeding 95\% accuracy. Notably, the AGS-GCN model achieved a perfect accuracy of 100\%, potentially attributed to the limited size of the dataset utilized in this study. Compared to the summary in Table \ref{tb_dl_speech}, for PD diagnosis and evaluation of gait data, researchers have focused on the feature extraction stage, employing simple global average pooling layers, fully connected layers, or Softmax layers on the classifier. Some deep learning works do not have a clear division between feature extraction and classification, opting for an end-to-end design. In such cases, Transformer-based models or LSTM-based models, along with some self-designed blocks, are used to gradually extract features and ultimately provide classification results, with the model performance also being good. For example, the FuseLGNet model performs exceptionally well on large sample data through convolutional blocks and FuseLGNet-specific blocks, achieving an accuracy of 99.78\%. Some studies in the table use the same dataset, and as observed through the performance metrics, the Transformer-based model achieves the best performance, while the CNN-based and LSTM-based models exhibit similar performance levels. Overall, employing a Transformer for feature extraction and classification may yield superior performance when applied to gait datasets.

\begin{table*}[!htp]
\centering 
\caption{Summarises the diagnosis and assessment of PD based on hand movement data using Deep learning techniques.}
\scalebox{0.8}{
\begin{tabular}{l|l|l|l|l|l|l|l}
  \hline
  \hline
 References&Dataset&No. of subjects&Feature extraction/selection&Classifier&Accuracy&Sensitivity&Specificity\\
\hline
\cite{9087433}&Signal&16-PD, 16-HC&Speed, Frequency, and Amplitude 
Estimates&KNN,\textbf{SVM},DT,and RF&98.4\%&*&*\\
\cite{9411721}&Signal&5-PD, 5-HC&Acceleration Signals Features&CNN&97.32\%&*&*\\
\cite{2022Two}&Video&12-PD, 12-HC&3D Motion Trajectory&Two-channel LSTM &95.7\%&95.8\%&92.8\%\\
\cite{ma2021retracted}&Signal&*&Slow Motion Features&SAG-RNN&93.55\%&*&*\\
\cite{guo2022tree}&Video&637-PD, 116-HC&Tri-directional Skeleton Tree Features& TSG-GCN &73.71\%&*&*\\
\cite{peng2024multi}&Signal&100-PD, 35-HC&Time, Frequency,Spectrum,and Autocorrelation&SVM,KNN,CNN,XGBoost,and \textbf{LGBM}&92.59\%&88\%&*\\
\cite{zhao2024selecting}&Signal&85-PD, 70-HC&Accelerometer Signals Features&\textbf{LGBM},SVM,KNN,and XGBoost&82.41\%&*&*\\
\cite{gazda2021multiple}&Image&68-PD, 73-HC&Spatial Features&CNN&94.7\%&*&*\\
\cite{wang2023coordinate}&Image&105-PD, 43-HC&Spatial Features&CAS Transformer&92.68\%&*&*\\
\cite{ma2022feature}&Image\&Signal&37-PD, 38-HC&Kinematic and CNN-based Features&DIT&98.87\%&*&*\\
\cite{CHERIET2023107344}&Video&20-PD, 23-HC&Skeleton Motion Featuress&Multi-speed Transformer&96.9\%&96.9\%&97.1\%\\
 \hline
 \hline
\end{tabular}}
  \label{tb_dl_hand}
\end{table*}

Traditional classifiers remained a practical tool for pattern recognition, utilizing the kinematic parameters of hand movements collected by a Leap Motion sensor for PD detection \cite{9087433}. Additionally, using wearable devices with inertial sensors to obtain acceleration, a CNN-based model was proposed to detect PD tremor \cite{9411721}. The 3D trajectory of hand motion with time series served PD severity rating \cite{2022Two}. Moreover, the processing of hand motion images and videos based on deep models was highly complex and contained a wealth of disease-related information, making it a focal point for researchers \cite{ma2021retracted,guo2022tree,sabapathy2022competent,peng2024multi,zhao2024selecting}. A tree-structure-guided graph convolutional network (TSG-GCN) \cite{guo2022tree} with contrastive learning was proposed to learn spatio-temporal features from 753 RGB video samples for the assessment of parkinsonian hand movements. CNNs, HMMs (Hidden Markov Models), and RNNs (Recurrent Neural Networks) also outperformed other deep models for rating hand movement disorders \cite{sabapathy2022competent,gazda2021multiple}. Methods based on self-attention mechanisms and Transformers were constantly emerging \cite{wang2023coordinate,ma2022feature,CHERIET2023107344}. Wang \textit{et al.} \cite{wang2023coordinate} proposed a Coordinate Attention Enhanced Swin Transformer (CAS Transformer) model that extracted the fuzzy edge features of handwriting images for PD detection.

Table \ref{tb_dl_hand} presents studies based on hand movement data using DL methods. The accuracy column represents the best result in the paper, and the corresponding optimal method has been highlighted. The results indicate that the accuracy of deep learning models based on hand movement data has reached a plateau. Aside from one case where data-related issues resulted in poor performance, other studies employing deep learning techniques on image data, sensor signal data, and video data achieved high levels of accuracy. Specifically, a study with a large number of participants achieved only a 73.71\% classification accuracy using a three-way skeleton tree feature extraction method and a TSG-GCN classifier on a dataset that included 637 Parkinson’s disease patients and 116 healthy controls. This lower accuracy may be due to the complex and chaotic nature of the dataset’s features, which may have introduced significant noise.In contrast, three algorithms achieved optimal performance on different types of hand movement data by adopting superior model frameworks (Transformer, multi-rate Transformer) or utilizing rich intrinsic data features. Among them, the DIT model learns the raw pen-tip kinematics data through a multi-head self-attention module and integrates multi-scale convolutional blocks in the embedding layer to extract helical features, combining the two types of motion features to achieve an accuracy of 98.87\%. Another study employed multiple classifiers and determined the optimal classifier (SVM), feature combination, and number of combined features by comparing different motion signal feature combinations. This study achieved an accuracy of 98.4\%. The last study is based on human skeletal features shown in videos and uses Multi-speed Transformer to capture joint sequences at different speeds, achieving the best performance of 96.9\% with hand video data. 

These methods all focus on one biological feature, one symptom or one modality of data, but there is still a lack of comprehensive consideration for the performance and development of PD.

\subsection{Single-task-based Research on PD} 
This section investigates the impact of single-task-based (PD detection/Severity Rating/Disease Tracking) gait, hand movements, and speech performance on the diagnosis and treatment of Parkinson's disease.

\begin{table*}[!htp]
\centering 
\caption{Single task of PD based on Multimodal data.}
\scalebox{0.8}{
\begin{tabular}{l|l|l|l|p{2.8cm}|l|l|l}
  \hline
  \hline
 References&Dataset&No. of subjects&Feature&Classifier&Accuracy&Sensitivity&Specificity\\
\hline
\cite{SADHU2022100351}&Hand Signal&4-PD, 5-HC&Speed,Amplitude,Hesitation,Halts,and Decrementing Amplitude&KNN,\textbf{SVM},DT&93\%&*&*\\
\cite{electronics8080907}&Hand Signal&62-PD, 15-HC&Coordinates, Pressure,and Grip Angle.&CNN&96.5\%&*&*\\
\cite{9376702}&Gait Signal&93-PD, 73-HC&Spatio-temporal Gait Features&HMM&98.93\%&*&*\\
\cite{9359526}&Gait Signal&93-PD, 73-HC&Spatio-temporal Gait Features&ASTCapsNet&97.31\%&*&*\\
\cite{9719945}&Handwriting \& Voice&81-PD, 85-HC&Spatial Features and MFCC&\textbf{DeiT},AST&92.37\%&*&*\\
\cite{MA2021102849}&Speech&34-PD, 20-HC&Original Features and Deep Features&Deep dual-side learning ensemble model&99.67\%&99.35\%&99.7\%\\
 \hline
 \hline
\end{tabular}}
  \label{tb_st}
\end{table*}

A machine learning-based telehealth infrastructure was proposed for remote symptom assessment, capable of detecting and classifying hand movement tasks based on the UPDRS \cite{SADHU2022100351}. For different hand movement tests in the UPDRS, CNNs were also utilized to detect and monitor PD from drawing movements \cite{electronics8080907}. Due to the prevalence of gait recognition algorithms, fusion networks had gradually been adopted for PD detection. A novel hybrid model was proposed to learn the gait differences between neurodegenerative diseases by fusing and aggregating data from multiple sensors \cite{9376702}. This model was capable of capturing temporal and structural information of bimodal gait data, i.e., images or signals. The vertical ground reaction force (VGRF, in Newton) was also input into an associated spatio-temporal capsule network for the severity rating of Parkinson's disease \cite{9359526}. A number of hybrid models \cite{9719945,MA2021102849,Zhao2023} had gradually emerged in the application of PD speech recognition. Specifically, an audio spectrogram transformer \cite{9719945} was proposed to analyze multimodal PD speech and handwriting data. An ensemble model \cite{MA2021102849} was designed for the classification of PD speech data, combining a deep sample learning algorithm with a deep network, realizing deep dual-side learning.

Compared to Table \ref{tb_dl_hand}, the evaluation results of single-task models using multimodal data in Table \ref{tb_st} show a certain degree of advancement and a clear trend toward stability. Notably, one study developed an automated learning system and trained a relevant spatiotemporal capsule network (ASTCapsNet) on a multisensor dataset to analyze multimodal information for gait recognition. The system designed a low-level feature extractor based on recurrent memory units and relational layers, and a high-level feature extractor for spatiotemporal feature extraction of gait. Subsequently, a Bayesian model was used to make decisions on class labels, achieving an accuracy of 97.31\% on a PD-related dataset. Additionally, based on spatiotemporal features, a study proposed a feature aggregation method for analyzing gait differences using data from multiple sensors. This approach utilized an SFE extractor to generate representative features from images or signals and designed a new Correlation Memory Neural Network (CorrMNN) structure to extract temporal features from two modalities, capturing temporal information. Finally, an HMM switch was embedded to associate observations with single state estimates, achieving an accuracy of 98.93\%. Based on multimodal speech data, another study further developed a deep dual-side learning ensemble model. This model performed deep feature learning using an embedded stack group sparse autoencoder and integrated these features with original speech features to create mixed feature data. A classification model was then trained on this feature space, achieving the highest accuracy of 99.67\%, making it one of the most accurate algorithms in related research.

The methods and datasets mentioned have indeed played a crucial role in advancing the diagnosis and treatment of Parkinson's disease (PD). However, studies that focus on single modalities and single tasks may encounter limitations in terms of applicability and flexibility. They might not fully accommodate the complexity and variability of the disease across different stages and settings. Additionally, their scalability might be limited, which could hinder their ability to encompass the intricate details of PD in diverse patient populations. To address these limitations, there is a need for more comprehensive, multimodal, and scalable approaches that can provide a more nuanced understanding of the disease.

\subsection{Multimodal Learning for PD Symptom} 
Deep learning methodologies play a major role in multimodal and multi-symptom recognition for Parkinson's Disease, significantly enhancing the diagnostic process by leveraging the rich information encapsulated within diverse data types. 

Deep learning algorithms adeptly learn temporal, spatial, and structural information from gait, hand movements, and speech in PD patients, integrating this multimodal information to provide diagnostic results. For instance, a hierarchical architecture \cite{wang2021hierarchical} combined Hidden Markov Models (HMMs), three single-symptom models, and machine learning to quantify multi-symptom severity in PD patients. A capsule network \cite{9359526} was trained on multimodal PD gait data to extract spatiotemporal and structural information. CNNs, LSTMs, ResNets, and Transformer-based hybrid models have been widely applied in multimodal PD symptom assessment \cite{junaid2023explainable,pahuja2022deep,CHERIET2023107344,li2023multimodal,faiem2024assessment,wang2024prediction,zhao2023spatio,xue2024ai}. A Spatio-Temporal Siamese Neural Network \cite{zhao2023spatio} was proposed for multimodal handwriting recognition, based on images and signals, and included a Siamese bidirectional memory neural network (SiamBiMNN) and a Siamese octave convolutional neural network (SiamOctCNN). A Perceiver architecture-based multimodal deep learning framework \cite{faiem2024assessment} was presented for analyzing gait time-series signals and hand-crafted features from GRF sensors, allowing them to complement each other and improve the predictive ability for PD diagnosis and PD symptom severity estimation. Furthermore, a multi-speed transformer architecture \cite{CHERIET2023107344} within a Deep Learning and Shallow Learning framework served for neurodegenerative disease classification and activity recognition.

In summary, the diagnostic and evaluative algorithms for Parkinson's disease are undergoing a significant transformation. The focus is shifting from machine learning to the more profound insights of deep learning, from the guidance of supervised models to the autonomy of unsupervised ones. The learning paradigm is expanding from single-modal to the rich integration of multimodal approaches, broadening the understanding of the disease's manifestations. It is also transitioning from tackling one task at a time to handling multiple tasks concurrently, enhancing the algorithms' versatility. Furthermore, the move from a single model to an ensemble of models is aimed at achieving a more robust, accurate, and dependable diagnostic process, ensuring that the evaluation results are both exhaustive and credible.

\begin{figure}[ht]
  \centering
  \includegraphics[width=8.5cm]{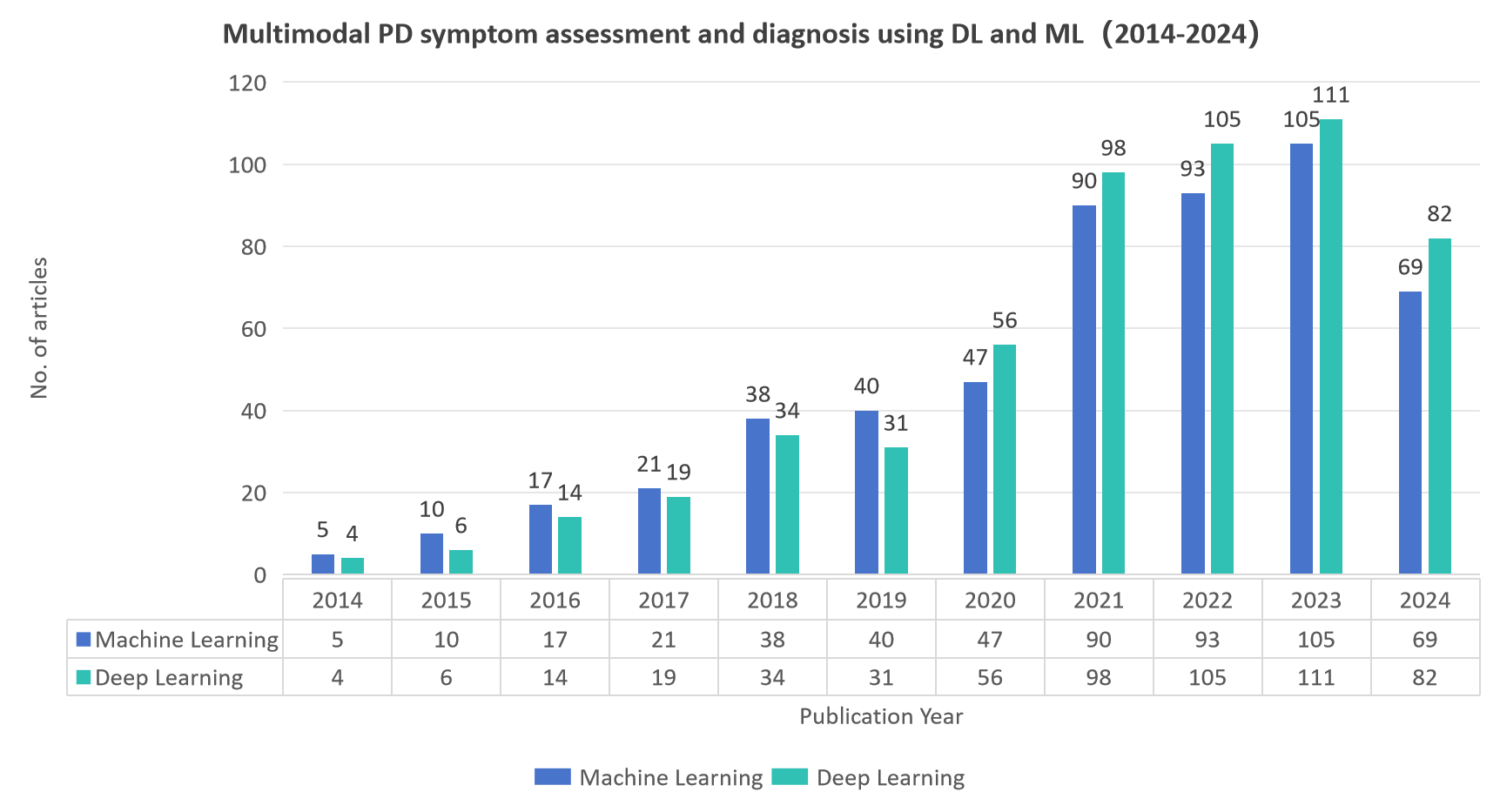}
  \caption{Multimodal PD symptom assessment and diagnosis methods using DL and ML (2014-2024).}
  \label{fig_dlmlpaper}
\end{figure}

Significantly, as shown in Fig. \ref{fig_dlmlpaper}, multimodal PD diagnosis and evaluation algorithms require significantly fewer learning iterations compared to their single-modal counterparts. Prior to 2020, the number of new single-modal machine learning algorithms increased by 5-40 per year. In contrast, the number of multimodal deep learning algorithms fluctuated between 4-31, generally maintaining a lower overall level than that of the machine learning algorithms. Post-2020, there was a surge; the number of both types of algorithms doubled, with deep learning taking the lead.

\begin{table*}[!htp]
\centering 
\caption{Multimodal PD symptom analysis using machine learning and deep learning techniques.}
\scalebox{0.9}{
\begin{tabular}{l|l|p{8cm}|l|l}
  \hline
  \hline
 References&Dataset&Modality&Method&Accuracy\\
\hline
\cite{li2024image}&Audio,motion and facial data&Image,signal&STN+RGA&95.08\%\\
\cite{bi2020multimodal}&fMRI+SNP&image&CERF&80.00\%\\
\cite{9376702}&Gait&VGRF+time series&CorrMNN&99.31\%\\
\cite{9359526}&Gait&Images, skeleton, force signal&ASTCapsNet&97.31\%\\
\cite{huo2020heterogeneous}&Bradykinesia, rigidity and tremor&Force signal,IMU,MMU&Voting Classifier&85.40\%\\
\cite{wang2021hierarchical}&Bradykinesia and tremor&Accelerometer and gyroscope data&HMMs&89.28\%\\
\cite{junaid2023explainable}&Motor and non-motor function data&Time-series,non timeseries&LGBM, RF&94.89\%\\
\cite{pahuja2022deep}&MRI and CSF&Neuroimaging and biological features&CNN&93.33\%\\
\cite{CHERIET2023107344}&Gait&Videos, time series&Multi-speed transformer network&96.90\%\\
\cite{li2023multimodal}&Gait&Inetial,stride, and pressure data&CNN-BiLSTM&98.89\%\\
\cite{faiem2024assessment}&Gait&GRF+time series&Cross-attention, Transformer&97.30\%\\
\cite{wang2024prediction}&Freezing of Gait&Acceleration of ankle,thigh,and trunk&MCT-Net&96.21\%\\
\cite{xue2024ai}&Multimodal-medical-data&Medical history, medication use, neuropsychological assessments, functional evaluations and multimodal neuroimaging&3D transformer-based Models&96\%(AUC)\\
 \hline
 \hline
\end{tabular}}
  \label{tb_multi-mldl}
\end{table*}

As shown in Table \ref{tb_multi-mldl}, we have introduced a total of 13 pieces of literature on multimodal PD detection and evaluation. This includes 6 works based on gait multimodal recognition, 2 works on MRI image processing and medical signal fusion, and 5 works on multi-symptom and multimodal fusion evaluation. These works target various data types such as images, videos, biological signals, physical signals, and temporal data, employing deep learning (DL) for feature extraction and training, and machine learning (ML) for classification and evaluation. Most DL methods are based on hybrid neural networks, with a few relying on ensemble learning and other techniques. The accuracy of most multimodal learning algorithms exceeds 90\%, and the performance of fusion models is generally superior to that of single models. The results of algorithms based on temporal networks and convolutional networks, as presented in the literature \cite{9376702, li2023multimodal}, outperform those in other studies. Due to their dependence on specific experimental conditions and datasets, other algorithms also exhibit outstanding advantages within their respective domains.




\section{Multi-Source Datasets}

The training and evaluation processes of learning models, irrespective of the underlying methodology—be it supervised, unsupervised, or any alternative approach—rely on datasets that encapsulate the essence of the task at hand. Utilizing these datasets not only facilitates the assessment of a method's efficacy in addressing particular challenges but also enables a comparative analysis with alternative strategies.

In this section, we delve into a variety of datasets tailored to the analysis of different motor and non-motor symptoms. These include speech, gait, hand movement, and multimodal datasets, each reflecting the complexity of Parkinson's disease. Varied in origin, format, composition, scale, and modality, these datasets offer a rich tapestry for the exploration and understanding of the disease's manifestations.

\subsection{Speech datasets}

\textbf{Oxford Parkinson's Disease Detection Dataset \cite{misc_parkinsons_174}:} This dataset is composed of a range of biomedical voice measurements from 31 individuals, 23 of whom have Parkinson's disease (PD). Each column in the table represents a specific voice measure, and each row corresponds to one of the 195 voice recordings from these individuals, as indicated in the "name" column. The primary objective of the data is to discriminate between healthy individuals and those with PD, based on the "status" column, which is set to 0 for healthy and 1 for PD.

\textbf{PDspeech Dataset \cite{misc_park301}:} This dataset consists of training and test files. The training data is derived from 20 people with Parkinson's (PWP), including 6 females and 14 males, and 20 healthy individuals, comprising 10 females and 10 males, who volunteered at the Department of Neurology in Cerrahpasa Faculty of Medicine, Istanbul University. From all subjects, multiple types of sound recordings were taken, including 26 voice samples with sustained vowels, numbers, words, and short sentences. A suite of 26 linear and time-frequency based features were extracted from each voice sample. The dataset also includes the UPDRS scores for each patient, as determined by an expert physician, allowing for its use in regression analysis.

\textbf{MDVR-KCL dataset \cite{hagen_jaeger_2019_2867216}:} This dataset comprises voice files of early and late Parkinson's disease patients and healthy controls, recorded with mobile devices at King's College London (KCL) Hospital in Brixton, London, from September 26 to 29, 2017. The recordings were made in a typical examination room with an area of about ten square meters and a typical reverberation time, utilizing a room with approximately ten square meters and a standard reverberation time for voice recording, with a duration of 500ms. The recording was conducted in a natural setting, with participants placing the phone to their preferred ear and positioning the microphone close to their mouth. It can be assumed that all recordings were made within the reverberation radius, suggesting they can be considered "clean."

A Motorola Moto G4 smartphone was used as the recording device. Through a developed application, high-quality recordings with a sampling rate of 44.1 kHz and a bit depth of 16 bits (CD audio quality) were achieved in '.wav' format. The dataset includes data from 16 PD patients and 21 healthy controls, with each participant's scores labeled on the Hoehn \& Yahr (H\&Y) scale, as well as the UPDRS II part 5 and UPDRS III part 18 scales.

\textbf{IPVS dataset \cite{aw6b-tg17-19}:} This dataset includes voice recordings from 28 PD patients and 20 healthy controls, all of which were collected at a 16 kHz sampling rate in a quiet, echo-free, and temperature-controlled room. The microphone was positioned 15 to 25 cm from the participants. The participants performed various tasks, including two phonations of the vowels /a/, /e/, /i/, /o/, and /u/, and syllable execution of "ka" and "pa" for 5 seconds. In this study, the reading of phonetically balanced phrases and vowel recordings were utilized, with a phonetically balanced text read twice by each participant.

\textbf{PDVoice dataset \cite{misc470}:} The data in this study were collected from 188 Parkinson's disease patients (107 males and 81 females) from the Neurology Department of Istanbul University, with ages ranging from 33 to 87 years old (mean age 65.1 ± 10.9). The control group consisted of 64 healthy individuals (23 males and 41 females), with ages ranging from 41 to 82 years old (mean age 61.1 ± 8.9). During the data collection process, the microphone was set to 44.1 kHz. After medical examination, sustained vocalizations of the vowel /a/ were collected from each subject, repeated three times.

\subsection{Hand-movement datasets} 
\begin{figure}
  \centering
  \includegraphics[width=8cm,height=5cm]{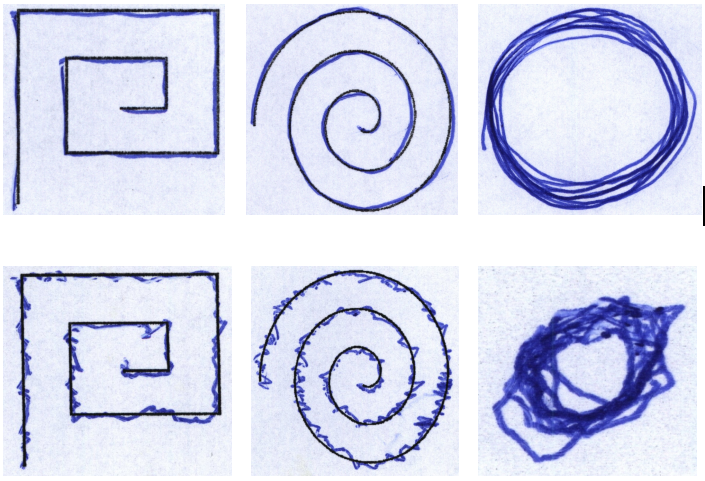}
  \caption{The handwriting data used in this project: We evaluate the proposed approach over the NewHandPD dataset, consists of the image data, i.e., spiral, meander, circled movements images.}
  \label{fig_handwriting}
\end{figure}

\begin{figure}[!htb]
  \centering
  \includegraphics[width=6cm]{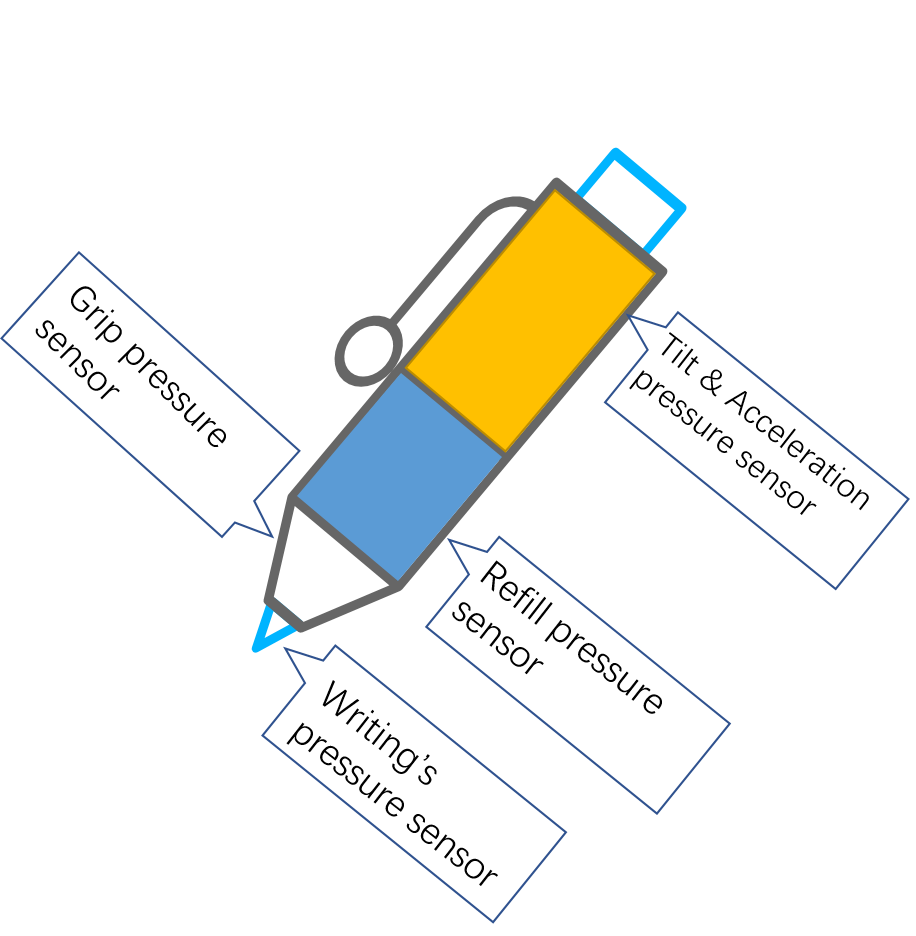}
  \caption{The biometric pen. It includes four sensors: tilt \& acceleration sensor, refill pressure sensor, grip pressure sensor, and writing's pressure sensor.}
  \label{pen}
\end{figure}

\begin{figure}[!htb]
  \centering
  \includegraphics[width=6cm]{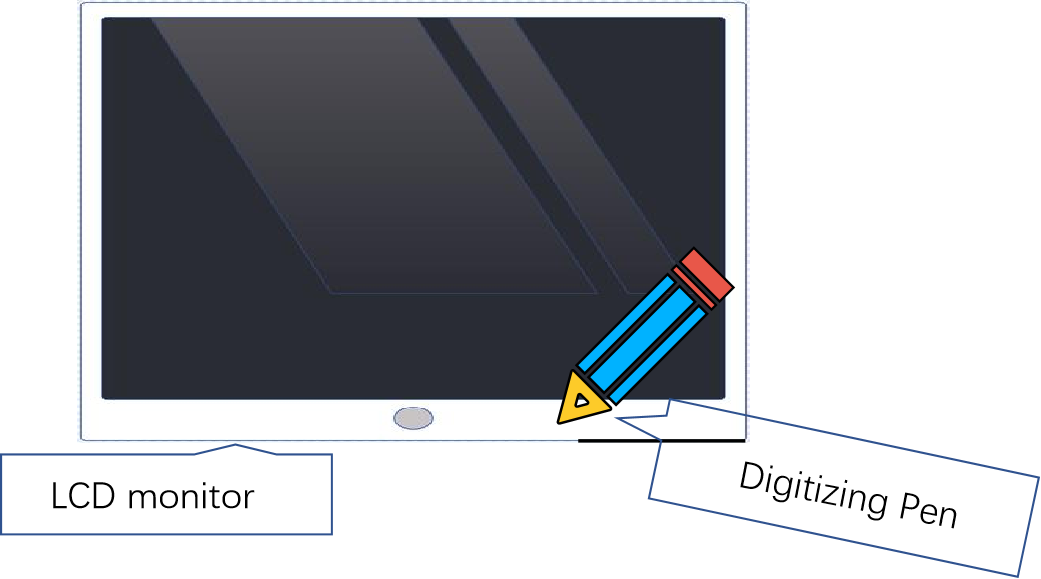}
  \caption{The digitized graphics tablet.}
  \label{tablet}
\end{figure}
\textbf{NewHandPD dataset \cite{7813053}:} This dataset includes images from two groups of individuals—healthy controls and PD patients—obtained during a handwriting test designed to assess their personal writing ability when filling out a form. The handwriting data was collected at the Botucatu Medical School of São Paulo State University, Brazil, and involved tasks unusual for PD patients, such as drawing "spirals," "meanders," and "circles" (Fig. \ref{fig_handwriting}). The dataset consisted of 66 participants (35 healthy controls and 31 PD patients), each asked to complete 12 tests, including four spirals, four meanders, and two circular movements (one in the air and the other on paper). After the test, each person had nine images saved. The dataset also includes signal data from a smart pen used in the diadochokinese test, where subjects performed hand-wrist movements while holding the pen with straight arms, and the generated signals were recorded. These signals were extracted from the BiSP$^\circledR$ smart pen (Fig. \ref{pen}), which has four sensors and six features, including microphone, fingergrip, axial pressure of the ink refill, and tilt and acceleration in the "X," "Y," and "Z" directions.

\textbf{PARKINSON\_HW dataset \cite{2014Improved}:} This dataset consists of 62 PD patients and 15 healthy controls who attended the Department of Neurology at Cerrahpasa Faculty of Medicine, Istanbul University. All participants underwent three handwriting tests: the Static Spiral Test, Dynamic Spiral Test, and Stability Test on Certain Point, using a Wacom Cintiq 12WX graphics tablet (Fig. \ref{tablet}). This tablet, integrated with an LCD monitor displaying the PC's screen, was used with specially designed software for recording handwritten graphics and testing coordination in PD patients. 

The Static Spiral Test (SST) was frequently used for clinical research, where patients were asked to trace three Archimedes spirals displayed on the drawing board using the software. The Dynamic Spiral Test (DST) differed by having the Archimedes spiral appear and disappear at certain intervals, challenging the patient to memorize and continue drawing the pattern. The Stability Test on Certain Point (STCP) aimed to measure the stability of the patient's hand or the degree of hand tremor, with the subject asked to hold a digital pen on a red dot in the middle of the screen.

\textbf{PD\_HW dataset \cite{misc_58}:} This PD and HC handwriting database comprises 25 individuals diagnosed with Parkinson's and 15 healthy individuals, all of whom visited the Department of Neurology at Cerrahpasa Faculty of Medicine, Istanbul University. A range of handwriting assessments was conducted, including the Static Spiral Test (SST), the Dynamic Spiral Test (DST), and the Stability Test on Certain Point (STCP). The dataset also contains images of the spirals drawn by the Parkinson's patients, suitable for image analysis, making the dataset valuable not only for handwriting analysis but also for regression analysis purposes.

\textbf{PD-Tappy Dataset \cite{10.1371/journal.pone.0188226}:} This PD and HC handwriting database comprises 25 individuals diagnosed with Parkinson's and 15 healthy individuals, all of whom visited the Department of Neurology at Cerrahpasa Faculty of Medicine, Istanbul University. A range of handwriting assessments was conducted, including the Static Spiral Test (SST), the Dynamic Spiral Test (DST), and the Stability Test on Certain Point (STCP). The dataset also contains images of the spirals drawn by the Parkinson's patients, suitable for image analysis, making the dataset valuable not only for handwriting analysis but also for regression analysis purposes.

\textbf{NeuroQWERTY MIT-CSXPD Dataset \cite{10.1038/srep34468}:} This dataset includes keystroke records from 85 individuals, both with and without PD. It demonstrates that regular computer keyboard use can help identify early motor symptoms of PD. Participants were recruited from two Madrid-based movement disorder units, following protocols approved by the Massachusetts Institute of Technology and Spanish hospitals. Using a Lenovo G50-70 laptop with Manjaro Linux, subjects typed on a standard word processor, simulating their home typing habits. The keystroke data, collected with a high temporal resolution, aimed to capture typing pattern nuances that could indicate early Parkinson's Disease symptoms, highlighting the potential of everyday computer interactions as a diagnostic tool for motor impairments in PD.

\textbf{PADS Dataset \cite{s41531-023-00625-7}:} These evaluations were recorded using a smart device-based system consisting of two smartwatches and one smartphone. The smartwatches, worn on the patient's wrists, synchronously recorded 11 interactive exercise tasks designed by neurologists. A total of 5159 measurement steps involving 469 individuals were captured. The PADS dataset includes all acceleration and rotation sensor signals, along with detailed information on movement steps, demographics, medical history, and PD-specific non-movement symptoms. We believe this extensively annotated dataset provides a suitable foundation for the training, validation, and optimization of future motion analysis technologies and sensor-driven systems. Three participant groups were recorded: 1) Parkinson's disease patients, 2) Differential diagnosis (DD) including primary tremor, atypical Parkinson's disease, secondary causes of Parkinson's disease, and multiple sclerosis, and 3) Healthy control group (HC).

\subsection{Gait dataset:} 

\textbf{PDgait dataset \cite{physionet-PDgait}:} This study utilized gait signals from PhysioNet, which consists of three PD gait sub-datasets contributed by three researchers. Ninety-three patients with idiopathic PD and 73 healthy controls (average age 66.3, 55\% male) were asked to walk at their usual, self-selected pace for about two minutes while wearing shoes equipped with 8 force sensors under each foot. The Vertical Ground Reaction Force (VGRF, in Newtons) was recorded by the force sensors as a function of time at 100 samples per second.

Each sample had 16 sensor outputs and two total outputs for each foot, with 19 parameters corresponding to each row in the dataset file: the timestamp, 16 VGRF foot parameters, and 2 total VGRF outputs. When subjects stood still, the 16 sensors recorded an initial value that changed as they began walking. With VGRF data, one can analyze the force record as a function of time and location, derive measures reflecting the center-of-pressure over time, and calculate stride or swing time for each foot. This allows for the study of stride-to-stride dynamics and the variability of these time series.

\textbf{NDDs dataset \cite{physionet-ndd}}: This dual-modal dataset contains temporal and force-sensing gait data from 48 patients with neurodegenerative diseases (ALS, HD, and PD) and 16 healthy controls (CO). The data were collected simultaneously using different sensors in a foot-switch system, which provided accurate estimates of the start and end stages of successive steps. The system utilized commercially available transducers and can be easily reproduced in laboratory environments. It included two 1.5 $in^2$ force sensing resistors and a 390$\Omega$ measuring resistor to acquire stride time intervals by measuring the variation in force changes during gait.

\textbf{tDCS FOG dataset \cite{Salomon2024article}}: The tDCS FOG data were originally collected as part of a sham-controlled, double-blind, multi-site randomized trial examining the effects of transcranial direct current stimulation on freezing of gait (FOG). The competition dataset included 71 PD patients with mild-to-moderate symptoms (Hoehn and Yahr score of 1-3.570) and self-reported FOG (via the NFOG-Q), who performed videotaped FOG-provoking tests while wearing a sensor with a 3D accelerometer placed on the lower back. Patients were excluded from the original study if they could not ambulate 20m unassisted, suffered from specific medical conditions detailed in the original work, or were unfit for the treatments. The FOG-provoking test protocol was conducted in the on-medication state and, if the patient agreed, also in the off-medication state. From the 71 tDCS FOG patients with three FOG class labels, 833 data series from 62 patients were included in the training data, 138 data series from 14 patients in the public test set, and 68 data series from 7 patients in the private test set.

\textbf{PD-FOG dataset \cite{misc_daphnet_245}}: This dataset contains annotated readings from 3 acceleration sensors at the hip and leg of Parkinson's disease patients experiencing freezing of gait (FoG) during walking tasks. The Daphnet Freezing of Gait Dataset was devised to benchmark automatic methods for recognizing gait freeze from wearable acceleration sensors placed on legs and the hip. The dataset was recorded in a lab with a focus on capturing numerous freeze events. Participants performed various tasks: straight line walking, walking with turns, and a realistic activity of daily living (ADL) task, such as going into different rooms to fetch coffee and opening doors.

\textbf{PD-Video dataset \cite{CHERIET2023107344}}: The dataset comprises 115 videos (61 of control subjects and 54 of people with diseases) from 43 subjects (22 females and 21 males, including 23 control subjects and 20 patients). The subjects walked a 4-meter straight line on the floor. The camera was positioned perpendicularly about 4 meters away from the line at a height of 2 meters. The videos varied in length and were recorded in different settings with various backgrounds and scenes. Each video shows a person walking a linear path in both directions, from left to right and vice versa, recorded at 25 fps. Control subjects ranged from 30-75 years, while patients were between 65-90 years. With the assistance of trained neurologists and psychologists, the disease stage was assessed using the Mini-Mental State Examination, classifying each subject as normal, mild, or severe.

\subsection{Multimodal dataset:} 

\textbf{PD-Posture-Gait dataset \cite{physionet-multimodal}}: This dataset offers a multi-camera, multimodal dataset for vision-based applications, utilizing a wheeled robotic walker equipped with a pair of affordable cameras. Depth data were acquired at 30 fps from a total of 14 healthy participants walking at three different gait speeds, across three different walking scenarios/paths at three different locations. Concurrently, accurate skeleton joint data were recorded using an inertial-based commercial motion capture system, providing a reliable ground-truth for both classical and novel (e.g., machine learning-based) vision-based applications. In total, the database contains approximately 166K frames of synchronized data, equating to 92 minutes of total recording time. This dataset may contribute to the development and evaluation of: i) classic or data-driven vision-based pose estimation algorithms; ii) applications in human detection and tracking, and movement forecasting; iii) and gait/posture metrics analysis using a rehabilitation device.

\textbf{REMAP dataset \cite{morgan2023multimodal}}:  This study recruited 24 participants, 12 with PD (mean age 61.25; 7 males, 5 females) and 12 healthy control volunteers (mean age 59.25, 3 males, 9 females). The controlled dataset contains pseudonymous data, including full skeleton data for all Sit-to-Stand (STS) and turning episodes. It also contains bilateral wrist-worn accelerometry data for these episodes from all participants, along with accelerometry data for non-turning, non-STS action episodes. The reason for including accelerometry data in the controlled dataset is due to the proven re-identifiability risk of accelerometry, which contravenes the study consent conditions relating to data sharing. We include individual-level demographic and clinical rating scale score outcomes given in ranges. The open dataset contains anonymous data, either because it is provided at the cohort-level (e.g., demographic data).

\textbf{Multimodal-FOG dataset \cite{li2021multimodal}}: This study gathered and presented a new multimodal dataset by combining rich physical and physiological sensor information. The multimodal data, including electroencephalogram (EEG), electromyogram (EMG), electrocardiogram (ECG), skin conductance (SC), and acceleration (ACC) in walking tasks, were collected using a high-quality hardware system with integrated commercial and self-designed sensors. A standard experimental procedure was carefully designed to induce FOG in hospital surroundings. A total of 12 PD patients completed the experiments, producing a total of 3 hours and 42 minutes of valid data. The FOG episodes in the multimodal data were labeled by two qualified physicians.

\textbf{Multi-PD-FOG dataset \cite{ribeiro2022public}}: This database includes videos, inertial measurement unit data, and clinical scales of freezing of gait in individuals with Parkinson's disease during a turning-in-place task. It encompasses both (a) patients' demography and (b) clinical conditions (PD severity, the number and duration of FOG episodes for each individual, clinical scales, and medication state during testing), as well as kinematics (video, acceleration, and angular velocity) during a turning-in-place task in individuals with PD in the ON medication state. A convenience sample of 35 idiopathic PD patients with FOG (16 females and 19 males) was recruited to participate in this study. The inertial sensors recorded triaxial linear accelerations and triaxial angular velocities at 128 Hz. Turning trials were recorded via a commercial digital camera (Sony, 30 Hz).

\textbf{Multi-Symptom dataset \cite{li2024image}}: This database includes videos, inertial measurement unit data, and clinical scales of freezing of gait in individuals with Parkinson's disease during a turning-in-place task. It encompasses both (a) patients' demography and (b) clinical conditions (PD severity, the number and duration of FOG episodes for each individual, clinical scales, and medication state during testing), as well as kinematics (video, acceleration, and angular velocity) during a turning-in-place task in individuals with PD in the ON medication state. A convenience sample of 35 idiopathic PD patients with FOG (16 females and 19 males) was recruited to participate in this study. The inertial sensors recorded triaxial linear accelerations and triaxial angular velocities at 128 Hz. Turning trials were recorded via a commercial digital camera (Sony, 30 Hz).

\textbf{Multimodal-medical dataset \cite{xue2024ai}}: This dataset includes individual-level demographics, health history, neurological testing, physical/neurological exams, and multisequence MRI scans. These data sources, when available, were aggregated from nine independent cohorts: 4RTNI, ADNI, AIBL, FHS, LBDSU, NACC, NIFD, OASIS, and PPMI. It collected demographics, personal and family history, laboratory results, findings from physical/neurological exams, medications, neuropsychological tests, and functional assessments, as well as multisequence magnetic resonance imaging (MRI) scans from 9 distinct cohorts, totaling 51,269 participants. All participants or their designated informants provided written informed consent. All protocols received approval from the respective institutional ethical review boards of each cohort. There were 19,849 participants with normal cognition (NC), 9,357 participants with mild cognitive impairment (MCI), and 22,063 participants with dementia.

\begin{figure}[!htb]
  \centering
  \includegraphics[width=8cm]{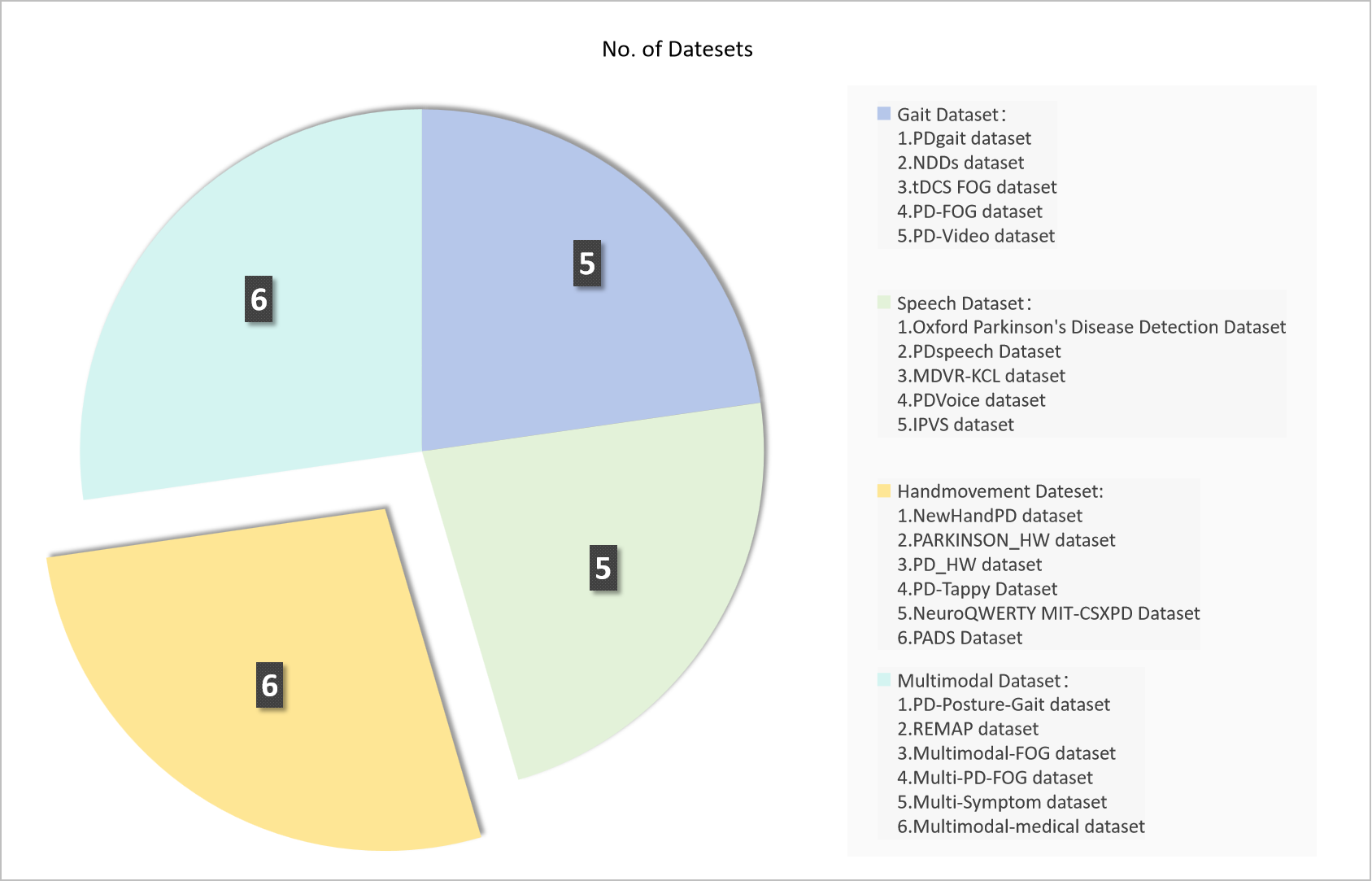}
  \caption{No. of the datasets introduced in this paper.}
  \label{datasetNo}
\end{figure}

\begin{figure}[!htb]
  \centering
  \includegraphics[width=8cm]{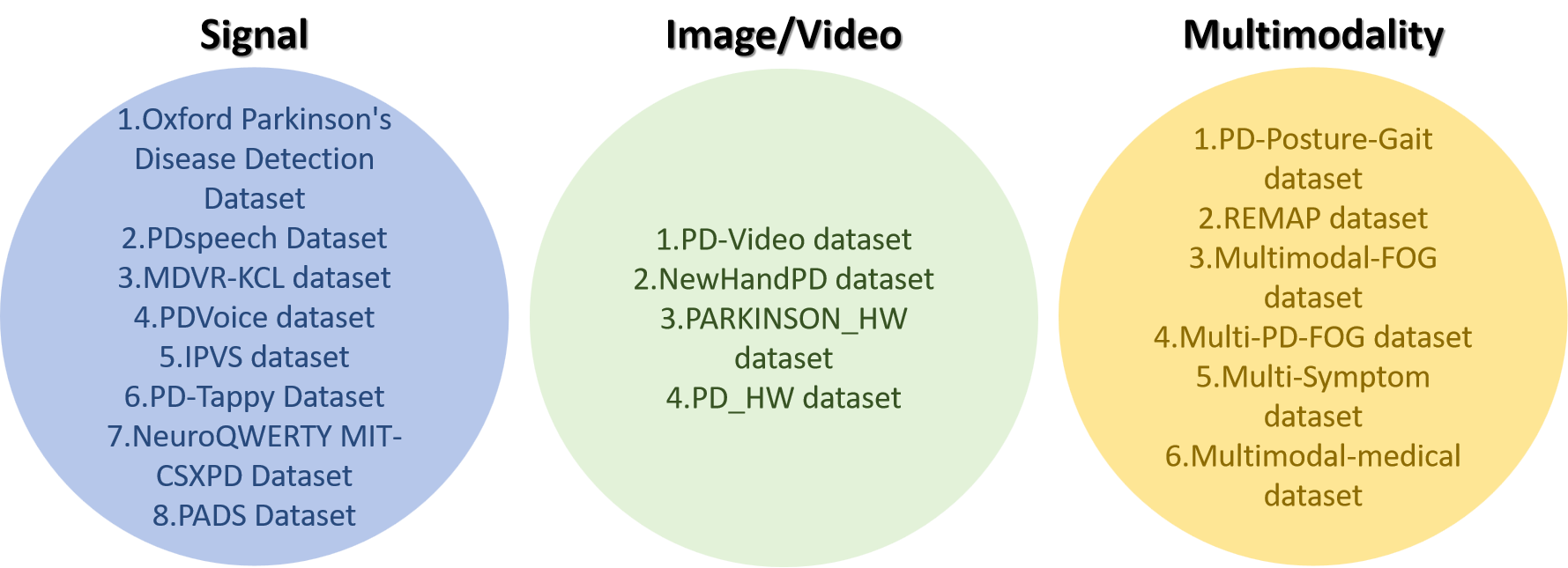}
  \caption{Partition of datasets with different modalities.}
  \label{datasetmodality}
\end{figure}

Fig. \ref{datasetNo} illustrates our presentation of four types of Parkinson's disease (PD) datasets, which include five speech datasets, six gait datasets, six hand action datasets, and five multimodal datasets. These are predominantly composed of data from three modalities: images, videos, and signals. Despite variations in signal sources, this paper does not further categorize them into subtypes such as physical or biological signals. Interested readers are encouraged to conduct detailed research using the dataset addresses provided in this article.

As depicted in Fig. \ref{datasetmodality}, the datasets are categorized into three groups based on their modality. There are 12 datasets that consist of signal data, while four datasets comprise image and video data. This distribution suggests that sensor-based data collection is more prevalent for Parkinson's disease patients, whereas image and video data collection faces challenges due to privacy concerns, as well as issues of redundancy and noise that necessitate further preprocessing. The limited image data of Parkinson's disease patients in multimodal datasets underscores the need for more diverse non-contact data types, such as images and videos, to foster new advancements in the assessment of PD's external symptoms.

\section{CONCLUSIONS AND FUTURE DIRECTIONS}
This article delivers an extensive review of machine learning and deep learning methodologies for the assessment and detection of Parkinson's disease symptoms spanning the period from 2014 to 2024. It encapsulates the profound effects of Parkinson's disease on three pivotal biological dimensions: speech, gait, and hand movements, elucidating the advancements in diagnostic accuracy and symptom recognition through the application of computational algorithms.

In the future, there will still be numerous challenges in this direction.

\textbf{i. Scarcity of Comprehensive Multimodal Datasets (multiple symptoms of PD):} Despite the availability of numerous datasets for Parkinson's disease focused on single symptoms, the absence of multimodal datasets encompassing a range of symptoms is evident. Diagnosing and treating Parkinson's disease by considering only one symptom is limiting and necessitates a more holistic approach with diverse PD data.

\textbf{ii. Wearable Device Impact:} The majority of PD data is collected through wearable sensors, which can influence patients' emotional states and psychological well-being during the data collection process. This not only demands considerable time and resources but also has the potential to compromise the reliability of the data. The integration of visual sensors, enhancing the role of video and image data in multimodal datasets, can help mitigate these shortcomings.

\textbf{iii. Algorithmic Constraints:} The algorithms discussed in this paper predominantly address single symptom and single modality data. While multimodal learning has emerged and gained popularity, the challenge of obtaining multimodal and multi-symptom PD data, coupled with the issue of heterogeneity, means that there is yet to be a medical framework for PD assessment and detection that is truly integrated into clinical practice.

\textbf{iv. Absence of Standardization:} Research outcomes vary widely due to the use of diverse sensors and data collection methods, leading to inconsistencies in data formats, scales, and conditions. This variability requires extensive data preprocessing and results in varying detection outcomes that can hinder diagnostic and treatment efficiency. Beyond the UPDRS, which serves as a reliable unified assessment tool, other standards are lacking. To tackle this issue, we advocate for the development of a novel evaluation paradigm.

\section*{Acknowledgment}
This research was supported in part by Natural Science Foundation of Shandong Province under Grant No.ZR2021QF084, National Natural Science Foundation of China under Grant No.62106117

\small
\bibliographystyle{ieeetr}
\bibliography{references}

\begin{thebibliography}{100}

\bibitem{YangArtificial2022}
Y.~Yuzhe, Y.~Yuan, Z.~Guo, W.~Hao, C.~Ying-Cong, L.~Yingcheng, T.~Christopher,
  G., C.~Daniel, B.~Jan, J.~Mithri, R., V.~Aleksandar, E.~Terry, D.,
  L.~Melissa, C., D.~Ray, and K.~Dina, ``Artificial intelligence-enabled
  detection and assessment of parkinson’s disease using nocturnal breathing
  signals,'' {\em Nature medicine}, vol.~10, no.~28, pp.~2207--2215, 2022.

\bibitem{Utilization2023}
F.~Meng, W.~Hu, S.~Wang, J.~Tam, Y.~Gao, X.~L. Zhu, D.~T.~M. Chan, W.~S. Poon,
  T.~L. Poon, F.~C. Cheung, B.~B.~T. Taw, L.~F. Li, S.~Y. Chen, K.~M. Chan,
  A.~Wang, Q.~Xu, C.~Han, Y.~Bai, A.~Wagle, Shukla, A.~Ramirez-Zamora,
  M.~Lozano, Andres, and J.~Zhang, ``Utilization, surgical populations,
  centers, coverages, regional balance, and their influential factors of deep
  brain stimulation for parkinson's disease: a large-scale multicenter
  cross-sectional study from 1997 to 2021,'' {\em International Journal of
  Surgery}, vol.~109, no.~11, pp.~3322--3336, 2023.

\bibitem{10.1007/978-3-030-59716-0_61}
M.~Lu, K.~Poston, A.~Pfefferbaum, E.~V. Sullivan, L.~Fei-Fei, K.~M. Pohl, J.~C.
  Niebles, and E.~Adeli, ``Vision-based estimation of mds-updrs gait scores for
  assessing parkinson's disease motor severity,'' in {\em Medical Image
  Computing and Computer Assisted Intervention -- MICCAI 2020} (A.~L. Martel,
  P.~Abolmaesumi, D.~Stoyanov, D.~Mateus, M.~A. Zuluaga, S.~K. Zhou,
  D.~Racoceanu, and L.~Joskowicz, eds.), (Cham), pp.~637--647, Springer
  International Publishing, 2020.

\bibitem{shahbakhi2014speech}
M.~Shahbakhi, D.~T. Far, and E.~Tahami, ``Speech analysis for diagnosis of
  parkinson’s disease using genetic algorithm and support vector machine,''
  {\em Journal of biomedical science and engineering}, vol.~2014, 2014.

\bibitem{benba2015voiceprints}
A.~Benba, A.~Jilbab, A.~Hammouch, and S.~Sandabad, ``Voiceprints analysis using
  mfcc and svm for detecting patients with parkinson's disease,'' in {\em 2015
  International conference on electrical and information technologies (ICEIT)},
  pp.~300--304, IEEE, 2015.

\bibitem{benba2015detecting}
A.~Benba, A.~Jilbab, and A.~Hammouch, ``Detecting patients with parkinson's
  disease using mel frequency cepstral coefficients and support vector
  machines,'' {\em International Journal on Electrical Engineering and
  Informatics}, vol.~7, no.~2, p.~297, 2015.

\bibitem{soumaya2019diagnosis}
Z.~Soumaya, B.~D. Taoufiq, B.~Nsiri, and A.~Abdelkrim, ``Diagnosis of parkinson
  disease using the wavelet transform and mfcc and svm classifier,'' in {\em
  2019 4th World Conference on Complex Systems (WCCS)}, pp.~1--6, IEEE, 2019.

\bibitem{solana2020automatic}
G.~Solana-Lavalle, J.-C. Gal{\'a}n-Hern{\'a}ndez, and R.~Rosas-Romero,
  ``Automatic parkinson disease detection at early stages as a pre-diagnosis
  tool by using classifiers and a small set of vocal features,'' {\em
  Biocybernetics and Biomedical Engineering}, vol.~40, no.~1, pp.~505--516,
  2020.

\bibitem{soumaya2021detection}
Z.~Soumaya, B.~D. Taoufiq, N.~Benayad, K.~Yunus, and A.~Abdelkrim, ``The
  detection of parkinson disease using the genetic algorithm and svm
  classifier,'' {\em Applied Acoustics}, vol.~171, p.~107528, 2021.

\bibitem{rahman2021parkinson}
A.~Rahman, S.~S. Rizvi, A.~Khan, A.~Afzaal~Abbasi, S.~U. Khan, and T.-S. Chung,
  ``Parkinson’s disease diagnosis in cepstral domain using mfcc and
  dimensionality reduction with svm classifier,'' {\em Mobile Information
  Systems}, vol.~2021, no.~1, p.~8822069, 2021.

\bibitem{9316078}
S.~Sharanyaa, P.~N. Renjith, and K.~Ramesh, ``Classification of parkinson's
  disease using speech attributes with parametric and nonparametric machine
  learning techniques,'' in {\em 2020 3rd International Conference on
  Intelligent Sustainable Systems (ICISS)}, pp.~437--442, 2020.

\bibitem{hawi2022automatic}
S.~Hawi, J.~Alhozami, R.~AlQahtani, D.~AlSafran, M.~Alqarni, and
  L.~El~Sahmarany, ``Automatic parkinson’s disease detection based on the
  combination of long-term acoustic features and mel frequency cepstral
  coefficients (mfcc),'' {\em Biomedical Signal Processing and Control},
  vol.~78, p.~104013, 2022.

\bibitem{indu2023modified}
R.~Indu, S.~C. Dimri, and P.~Malik, ``A modified knn algorithm to detect
  parkinson’s disease,'' {\em Network Modeling Analysis in Health Informatics
  and Bioinformatics}, vol.~12, no.~1, p.~24, 2023.

\bibitem{pramanik2021machine}
M.~Pramanik, R.~Pradhan, P.~Nandy, A.~K. Bhoi, and P.~Barsocchi, ``Machine
  learning methods with decision forests for parkinson’s detection,'' {\em
  Applied Sciences}, vol.~11, no.~2, p.~581, 2021.

\bibitem{ouhmida2022parkinson}
A.~Ouhmida, A.~Raihani, B.~Cherradi, and Y.~Lamalem, ``Parkinson's disease
  classification using machine learning algorithms: performance analysis and
  comparison,'' in {\em 2022 2nd international conference on innovative
  research in applied science, engineering and technology (IRASET)}, pp.~1--6,
  IEEE, 2022.

\bibitem{zhang2016classification}
H.-H. Zhang, L.~Yang, Y.~Liu, P.~Wang, J.~Yin, Y.~Li, M.~Qiu, X.~Zhu, and
  F.~Yan, ``Classification of parkinson’s disease utilizing multi-edit
  nearest-neighbor and ensemble learning algorithms with speech samples,'' {\em
  Biomedical engineering online}, vol.~15, pp.~1--22, 2016.

\bibitem{liu2021local}
Y.~Liu, Y.~Li, X.~Tan, P.~Wang, and Y.~Zhang, ``Local discriminant preservation
  projection embedded ensemble learning based dimensionality reduction of
  speech data of parkinson’s disease,'' {\em Biomedical Signal Processing and
  Control}, vol.~63, p.~102165, 2021.

\bibitem{barukab2022analysis}
O.~Barukab, A.~Ahmad, T.~Khan, and M.~R. Thayyil~Kunhumuhammed, ``Analysis of
  parkinson’s disease using an imbalanced-speech dataset by employing
  decision tree ensemble methods,'' {\em Diagnostics}, vol.~12, no.~12,
  p.~3000, 2022.

\bibitem{oung2018evaluation}
Q.~W. Oung, S.~N. Basah, H.~Muthusamy, V.~Vijean, and H.~Lee, ``Evaluation of
  short-term cepstral based features for detection of parkinson’s disease
  severity levels through speech signals,'' in {\em IOP Conference Series:
  Materials Science and Engineering}, vol.~318, pp.~012--039, IOP Publishing,
  2018.

\bibitem{kuresan2019fusion}
H.~Kuresan, D.~Samiappan, and S.~Masunda, ``Fusion of wpt and mfcc feature
  extraction in parkinson’s disease diagnosis,'' {\em Technology and Health
  Care}, vol.~27, no.~4, pp.~363--372, 2019.

\bibitem{boualoulou2023comparison}
N.~Boualoulou, T.~Belhoussine~Drissi, and B.~Nsiri, ``Comparison of feature
  extraction methods between mfcc, bfcc, and gfcc with svm classifier for
  parkinson’s disease diagnosis,'' in {\em International Conference on IoT
  Based Control Networks and Intelligent Systems}, pp.~231--247, Springer,
  2023.

\bibitem{nissar2019voice}
I.~Nissar, D.~R. Rizvi, S.~Masood, and A.~N. Mir, ``Voice-based detection of
  parkinson’s disease through ensemble machine learning approach: A
  performance study,'' {\em EAI endorsed transactions on pervasive health and
  technology}, vol.~5, no.~19, pp.~e2--e2, 2019.

\bibitem{9793048}
G.~Tallapureddy and D.~Radha, ``Analysis of ensemble of machine learning
  algorithms for detection of parkinson's disease,'' in {\em 2022 International
  Conference on Applied Artificial Intelligence and Computing (ICAAIC)},
  pp.~354--361, 2022.

\bibitem{shastry2023ensemble}
K.~A. Shastry, ``An ensemble nearest neighbor boosting technique for prediction
  of parkinson’s disease,'' {\em Healthcare Analytics}, vol.~3, p.~100181,
  2023.

\bibitem{celik2023proposing}
G.~Celik and E.~Ba{\c{s}}aran, ``Proposing a new approach based on
  convolutional neural networks and random forest for the diagnosis of
  parkinson's disease from speech signals,'' {\em Applied Acoustics}, vol.~211,
  p.~109476, 2023.

\bibitem{zlotnik2015random}
A.~Zlotnik, J.~M. Montero, R.~San-Segundo, and A.~Gallardo-Antol{\'\i}n,
  ``Random forest-based prediction of parkinson's disease progression using
  acoustic, asr and intelligibility features,'' in {\em Sixteenth Annual
  Conference of the International Speech Communication Association}, 2015.

\bibitem{kim2021abnormal}
W.~Kim and Y.~Kim, ``Abnormal gait recognition based on integrated gait
  features in machine learning,'' in {\em 2021 IEEE 45th Annual Computers,
  Software, and Applications Conference (COMPSAC)}, pp.~1683--1688, IEEE, 2021.

\bibitem{zhao2018hybrid}
A.~Zhao, L.~Qi, J.~Li, J.~Dong, and H.~Yu, ``A hybrid spatio-temporal model for
  detection and severity rating of parkinson’s disease from gait data,'' {\em
  Neurocomputing}, vol.~315, pp.~1--8, 2018.

\bibitem{khan2021novel}
T.~Khan, A.~Zeeshan, and M.~Dougherty, ``A novel method for automatic
  classification of parkinson gait severity using front-view video analysis,''
  {\em Technology and Health Care}, vol.~29, no.~4, pp.~643--653, 2021.

\bibitem{wahid2015classification}
F.~Wahid, R.~K. Begg, C.~J. Hass, S.~Halgamuge, and D.~C. Ackland,
  ``Classification of parkinson's disease gait using spatial-temporal gait
  features,'' {\em IEEE journal of biomedical and health informatics}, vol.~19,
  no.~6, pp.~1794--1802, 2015.

\bibitem{shetty2016svm}
S.~Shetty and Y.~Rao, ``Svm based machine learning approach to identify
  parkinson's disease using gait analysis,'' in {\em 2016 International
  conference on inventive computation technologies (ICICT)}, vol.~2, pp.~1--5,
  IEEE, 2016.

\bibitem{abdulhay2018gait}
E.~Abdulhay, N.~Arunkumar, K.~Narasimhan, E.~Vellaiappan, and V.~Venkatraman,
  ``Gait and tremor investigation using machine learning techniques for the
  diagnosis of parkinson disease,'' {\em Future Generation Computer Systems},
  vol.~83, pp.~366--373, 2018.

\bibitem{alkhatib2020machine}
R.~Alkhatib, M.~O. Diab, C.~Corbier, and M.~El~Badaoui, ``Machine learning
  algorithm for gait analysis and classification on early detection of
  parkinson,'' {\em IEEE Sensors Letters}, vol.~4, no.~6, pp.~1--4, 2020.

\bibitem{balaji2020supervised}
E.~Balaji, D.~Brindha, and R.~Balakrishnan, ``Supervised machine learning based
  gait classification system for early detection and stage classification of
  parkinson’s disease,'' {\em Applied Soft Computing}, vol.~94, p.~106494,
  2020.

\bibitem{chavez2022vision}
J.~M. Chavez and W.~Tang, ``A vision-based system for stage classification of
  parkinsonian gait using machine learning and synthetic data,'' {\em Sensors},
  vol.~22, no.~12, p.~4463, 2022.

\bibitem{chen2022computer}
B.~Chen, C.~Chen, J.~Hu, Z.~Sayeed, J.~Qi, H.~F. Darwiche, B.~E. Little,
  S.~Lou, M.~Darwish, C.~Foote, {\em et~al.}, ``Computer vision and machine
  learning-based gait pattern recognition for flat fall prediction,'' {\em
  Sensors}, vol.~22, no.~20, p.~7960, 2022.

\bibitem{loureiro2020using}
J.~Loureiro and P.~L. Correia, ``Using a skeleton gait energy image for
  pathological gait classification,'' in {\em 2020 15th IEEE International
  Conference on Automatic Face and Gesture Recognition (FG 2020)},
  pp.~503--507, IEEE, 2020.

\bibitem{munoz2022machine}
B.~Mu{\~n}oz-Ospina, D.~Alvarez-Garcia, H.~J.~C. Clavijo-Moran, J.~A.
  Valderrama-Chaparro, M.~Garc{\'\i}a-Pe{\~n}a, C.~A. Herr{\'a}n, C.~C.
  Urcuqui, A.~Navarro-Cadavid, and J.~Orozco, ``Machine learning classifiers to
  evaluate data from gait analysis with depth cameras in patients with
  parkinson’s disease,'' {\em Frontiers in Human Neuroscience}, vol.~16,
  p.~826376, 2022.

\bibitem{orphanidou2018predicting}
N.~K. Orphanidou, A.~Hussain, R.~Keight, P.~Lishoa, J.~Hind, and H.~Al-Askar,
  ``Predicting freezing of gait in parkinsons disease patients using machine
  learning,'' in {\em 2018 IEEE Congress on Evolutionary Computation (CEC)},
  pp.~1--8, IEEE, 2018.

\bibitem{gong2020novel}
L.~Gong, J.~Li, M.~Yu, M.~Zhu, and R.~Clifford, ``A novel computer vision based
  gait analysis technique for normal and parkinson’s gaits classification,''
  in {\em 2020 IEEE Intl Conf on Dependable, Autonomic and Secure Computing,
  Intl Conf on Pervasive Intelligence and Computing, Intl Conf on Cloud and Big
  Data Computing, Intl Conf on Cyber Science and Technology Congress
  (DASC/PiCom/CBDCom/CyberSciTech)}, pp.~209--215, IEEE, 2020.

\bibitem{borzi2023context}
L.~Borz{\`\i}, L.~Sigcha, and G.~Olmo, ``Context recognition algorithms for
  energy-efficient freezing-of-gait detection in parkinson’s disease,'' {\em
  Sensors}, vol.~23, no.~9, p.~4426, 2023.

\bibitem{kleanthous2020new}
N.~Kleanthous, A.~J. Hussain, W.~Khan, and P.~Liatsis, ``A new machine learning
  based approach to predict freezing of gait,'' {\em Pattern Recognition
  Letters}, vol.~140, pp.~119--126, 2020.

\bibitem{islam2024review}
M.~A. Islam, M.~Z.~H. Majumder, M.~A. Hussein, K.~M. Hossain, and M.~S. Miah,
  ``A review of machine learning and deep learning algorithms for parkinson's
  disease detection using handwriting and voice datasets,'' {\em Heliyon},
  2024.

\bibitem{impedovo2018dynamic}
D.~Impedovo, G.~Pirlo, and G.~Vessio, ``Dynamic handwriting analysis for
  supporting earlier parkinson’s disease diagnosis,'' {\em Information},
  vol.~9, no.~10, p.~247, 2018.

\bibitem{pereira2015step}
C.~R. Pereira, D.~R. Pereira, F.~A. Da~Silva, C.~Hook, S.~A. Weber, L.~A.
  Pereira, and J.~P. Papa, ``A step towards the automated diagnosis of
  parkinson's disease: Analyzing handwriting movements,'' in {\em 2015 IEEE
  28th international symposium on computer-based medical systems},
  pp.~171--176, Ieee, 2015.

\bibitem{ranjan2023detection}
N.~M. Ranjan, G.~Mate, and M.~Bembde, ``Detection of parkinson's disease using
  machine learning algorithms and handwriting analysis,'' {\em Journal of Data
  Mining and Management}, vol.~8, no.~1, pp.~21--29, 2023.

\bibitem{kotsavasiloglou2017machine}
C.~Kotsavasiloglou, N.~Kostikis, D.~Hristu-Varsakelis, and M.~Arnaoutoglou,
  ``Machine learning-based classification of simple drawing movements in
  parkinson's disease,'' {\em Biomedical Signal Processing and Control},
  vol.~31, pp.~174--180, 2017.

\bibitem{castrillon2019characterization}
R.~Castrill{\'o}n, A.~Acien, J.~R. Orozco-Arroyave, A.~Morales, J.~Vargas,
  R.~Vera-Rodr{\'\i}guez, J.~Fi{\'e}rrez, J.~Ortega-Garcia, and A.~Villegas,
  ``Characterization of the handwriting skills as a biomarker for parkinson’s
  disease,'' in {\em 2019 14th IEEE International Conference on Automatic Face
  \& Gesture Recognition (FG 2019)}, pp.~1--5, IEEE, 2019.

\bibitem{guarin2024characterizing}
D.~L. Guar{\'\i}n, J.~K. Wong, N.~R. McFarland, and A.~Ramirez-Zamora,
  ``Characterizing disease progression in parkinson’s disease from videos of
  the finger tapping test,'' {\em IEEE Transactions on Neural Systems and
  Rehabilitation Engineering}, 2024.

\bibitem{yang2022automatic}
N.~Yang, D.-F. Liu, T.~Liu, T.~Han, P.~Zhang, X.~Xu, S.~Lou, H.-G. Liu, A.-C.
  Yang, C.~Dong, {\em et~al.}, ``Automatic detection pipeline for accessing the
  motor severity of parkinson’s disease in finger tapping and postural
  stability,'' {\em IEEE Access}, vol.~10, pp.~66961--66973, 2022.

\bibitem{yu2023clinically}
T.~Yu, K.~W. Park, M.~J. McKeown, and Z.~J. Wang, ``Clinically informed
  automated assessment of finger tapping videos in parkinson’s disease,''
  {\em Sensors}, vol.~23, no.~22, p.~9149, 2023.

\bibitem{khan2014computer}
T.~Khan, D.~Nyholm, J.~Westin, and M.~Dougherty, ``A computer vision framework
  for finger-tapping evaluation in parkinson's disease,'' {\em Artificial
  intelligence in medicine}, vol.~60, no.~1, pp.~27--40, 2014.

\bibitem{he2023effective}
Q.~He and H.~Gao, ``An effective hand pose estimation based evaluation method
  in assessing parkinson’s finger tap movements,'' in {\em International
  Symposium on Artificial Intelligence and Robotics}, pp.~511--518, Springer,
  2023.

\bibitem{serrano2024estimation}
J.~I. Serrano, J.~P. Romero, A.~Arroyo-Ferrer, and M.~D. del Castillo,
  ``Estimation of motor severity scales in parkinson’s disease by linear
  models of bimanual non-alternating index finger tapping features,'' {\em
  Expert Systems with Applications}, vol.~246, p.~123077, 2024.

\bibitem{TODOROV2024146}
D.~Todorov, A.~Schnitzler, and J.~Hirschmann, ``Parkinsonian rest tremor can be
  distinguished from voluntary hand movements based on subthalamic and cortical
  activity,'' {\em Clinical Neurophysiology}, vol.~157, pp.~146--155, 2024.

\bibitem{nahiduzzaman2020machine}
M.~Nahiduzzaman, M.~Tasnim, N.~T. Newaz, M.~S. Kaiser, and M.~Mahmud, ``Machine
  learning based early fall detection for elderly people with neurological
  disorder using multimodal data fusion,'' in {\em International Conference on
  Brain Informatics}, pp.~204--214, Springer, 2020.

\bibitem{li2024image}
J.~Li, Y.~Zhao, H.~Zhang, W.~J. LiMember, C.~Fu, C.~Lian, and P.~Shan, ``Image
  encoding and fusion of multi-modal data enhance depression diagnosis in
  parkinson's disease patients,'' {\em IEEE Transactions on Affective
  Computing}, 2024.

\bibitem{bi2020multimodal}
X.-a. Bi, X.~Hu, H.~Wu, and Y.~Wang, ``Multimodal data analysis of alzheimer's
  disease based on clustering evolutionary random forest,'' {\em IEEE Journal
  of Biomedical and Health Informatics}, vol.~24, no.~10, pp.~2973--2983, 2020.

\bibitem{9376702}
A.~Zhao, J.~Li, J.~Dong, L.~Qi, Q.~Zhang, N.~Li, X.~Wang, and H.~Zhou,
  ``Multimodal gait recognition for neurodegenerative diseases,'' {\em IEEE
  Transactions on Cybernetics}, vol.~52, no.~9, pp.~9439--9453, 2022.

\bibitem{10546910}
R.~M. Al-Tam, F.~A. Hashim, S.~Maqsood, L.~Abualigah, and R.~M. Alwhaibi,
  ``Enhancing parkinson’s disease diagnosis through stacking ensemble-based
  machine learning approach,'' {\em IEEE Access}, vol.~12, pp.~79549--79567,
  2024.

\bibitem{makarious2022multi}
M.~B. Makarious, H.~L. Leonard, D.~Vitale, H.~Iwaki, L.~Sargent, A.~Dadu,
  I.~Violich, E.~Hutchins, D.~Saffo, S.~Bandres-Ciga, {\em et~al.},
  ``Multi-modality machine learning predicting parkinson’s disease,'' {\em
  npj Parkinson's Disease}, vol.~8, no.~1, p.~35, 2022.

\bibitem{huo2020heterogeneous}
W.~Huo, P.~Angeles, Y.~F. Tai, N.~Pavese, S.~Wilson, M.~T. Hu, and
  R.~Vaidyanathan, ``A heterogeneous sensing suite for multisymptom
  quantification of parkinson’s disease,'' {\em IEEE Transactions on Neural
  Systems and Rehabilitation Engineering}, vol.~28, no.~6, pp.~1397--1406,
  2020.

\bibitem{asdadasda}
J.~C. V{\'a}squez-Correa, N.~Garcia-Ospina, J.~R. Orozco-Arroyave, M.~Cernak,
  and E.~N{\"o}th, ``Phonological posteriors and gru recurrent units to assess
  speech impairments of patients with parkinson's disease,'' in {\em Text,
  Speech, and Dialogue} (P.~Sojka, A.~Hor{\'a}k, I.~Kope{\v{c}}ek, and K.~Pala,
  eds.), pp.~453--461, Springer International Publishing, 2018.

\bibitem{Hernandez2022CrosslingualSS}
A.~Hernandez, P.~A. P'erez-Toro, E.~N{\"o}th, J.~R. Orozco-Arroyave, A.~K.
  Maier, and S.~H. Yang, ``Cross-lingual self-supervised speech representations
  for improved dysarthric speech recognition,'' {\em ArXiv},
  vol.~abs/2204.01670, 2022.

\bibitem{bhati2019lstm}
S.~Bhati, L.~M. Velazquez, J.~Villalba, and N.~Dehak, ``Lstm siamese network
  for parkinson’s disease detection from speech,'' in {\em 2019 ieee global
  conference on signal and information processing (globalsip)}, pp.~1--5, IEEE,
  2019.

\bibitem{gunduz2019deep}
H.~Gunduz, ``Deep learning-based parkinson’s disease classification using
  vocal feature sets,'' {\em Ieee access}, vol.~7, pp.~115540--115551, 2019.

\bibitem{vasquez2017convolutional}
J.~C. V{\'a}squez-Correa, J.~R. Orozco-Arroyave, and E.~N{\"o}th,
  ``Convolutional neural network to model articulation impairments in patients
  with parkinson's disease.,'' in {\em Interspeech}, pp.~314--318, Stockholm,
  2017.

\bibitem{zhang2018deepvoice}
H.~Zhang, A.~Wang, D.~Li, and W.~Xu, ``Deepvoice: A voiceprint-based mobile
  health framework for parkinson's disease identification,'' in {\em 2018 IEEE
  EMBS International Conference on Biomedical \& Health Informatics (BHI)},
  pp.~214--217, IEEE, 2018.

\bibitem{zhao2024triplet}
A.~Zhao, N.~Wang, X.~Niu, M.~Chen, and H.~Wu, ``A triplet multimodel transfer
  learning network for speech disorder screening of parkinson’s disease,''
  {\em International Journal of Intelligent Systems}, vol.~2024, no.~1,
  p.~8890592, 2024.

\bibitem{hemmerling2023vision}
D.~Hemmerling, M.~Wodzinski, J.~R. Orozco-Arroyave, D.~Sztaho, M.~Daniol,
  P.~Jemiolo, and M.~Wojcik-Pedziwiatr, ``Vision transformer for parkinson’s
  disease classification using multilingual sustained vowel recordings,'' in
  {\em 2023 45th Annual International Conference of the IEEE Engineering in
  Medicine \& Biology Society (EMBC)}, pp.~1--4, IEEE, 2023.

\bibitem{nijhawan2023novel}
R.~Nijhawan, M.~Kumar, S.~Arya, N.~Mendirtta, S.~Kumar, S.~Towfek, D.~S.
  Khafaga, H.~K. Alkahtani, and A.~A. Abdelhamid, ``A novel
  artificial-intelligence-based approach for classification of parkinson’s
  disease using complex and large vocal features,'' {\em Biomimetics}, vol.~8,
  no.~4, p.~351, 2023.

\bibitem{fang2020parkinsonian}
H.~Fang, C.~Gong, C.~Zhang, Y.~Sui, and L.~Li, ``Parkinsonian chinese speech
  analysis towards automatic classification of parkinson's disease,'' in {\em
  Machine Learning for Health}, pp.~114--125, PMLR, 2020.

\bibitem{mehra2024deep}
S.~Mehra, V.~Ranga, and R.~Agarwal, ``A deep learning approach to dysarthric
  utterance classification with bilstm-gru, speech cue filtering, and log mel
  spectrograms,'' {\em The Journal of Supercomputing}, pp.~1--28, 2024.

\bibitem{jeong2024exploring}
S.-M. Jeong, S.~Kim, E.~C. Lee, and H.~J. Kim, ``Exploring spectrogram-based
  audio classification for parkinson’s disease: A study on speech
  classification and qualitative reliability verification,'' {\em Sensors},
  vol.~24, no.~14, p.~4625, 2024.

\bibitem{klempivr2023evaluating}
O.~Klemp{\'\i}{\v{r}}, D.~P{\v{r}}{\'\i}hoda, and R.~Krupi{\v{c}}ka,
  ``Evaluating the performance of wav2vec embedding for parkinson's disease
  detection,'' {\em Measurement Science Review}, vol.~23, no.~6, pp.~260--267,
  2023.

\bibitem{9850832}
M.~Dhakal, A.~Chhetri, A.~K. Gupta, P.~Lamichhane, S.~Pandey, and S.~Shakya,
  ``Automatic speech recognition for the nepali language using cnn,
  bidirectional lstm and resnet,'' in {\em 2022 International Conference on
  Inventive Computation Technologies (ICICT)}, pp.~515--521, 2022.

\bibitem{van2024innovative}
L.~van Gelderen and C.~Tejedor-Garc{\'\i}a, ``Innovative speech-based deep
  learning approaches for parkinson's disease classification: A systematic
  review,'' {\em arXiv preprint arXiv:2407.17844}, 2024.

\bibitem{klempir2024analyzing1}
O.~Klempir and R.~Krupicka, ``Analyzing wav2vec embedding in parkinson's
  disease speech: A study on cross-database classification and regression
  tasks,'' {\em medRxiv}, pp.~2024--04, 2024.

\bibitem{chowdary2023few}
P.~N. Chowdary, M.~S. Akshay, V.~S. Aravind, M.~S. Aashish, G.~V.~V. Vardhan,
  and G.~J. Lal, ``A few-shot approach to dysarthric speech intelligibility
  level classification using transformers,'' in {\em 2023 14th International
  Conference on Computing Communication and Networking Technologies (ICCCNT)},
  pp.~1--6, IEEE, 2023.

\bibitem{jiang2023self}
H.~Jiang, ``Self-supervised learning for early detection of neurodegenerative
  diseases with small data,'' {\em Doctoral thesis}, 2023.

\bibitem{zhou2023risevi}
F.~Zhou, S.~Hu, X.~Wan, Z.~Lu, and J.~Wu, ``Risevi: A disease risk prediction
  model based on vision transformer applied to nursing homes,'' {\em
  Electronics}, vol.~12, no.~15, p.~3206, 2023.

\bibitem{ZHAO201891}
A.~Zhao, L.~Qi, J.~Dong, and H.~Yu, ``Dual channel lstm based multi-feature
  extraction in gait for diagnosis of neurodegenerative diseases,'' {\em
  Knowledge-Based Systems}, vol.~145, pp.~91--97, 2018.

\bibitem{TIAN2022116}
H.~Tian, X.~Ma, H.~Wu, and Y.~Li, ``Skeleton-based abnormal gait recognition
  with spatio-temporal attention enhanced gait-structural graph convolutional
  networks,'' {\em Neurocomputing}, vol.~473, pp.~116--126, 2022.

\bibitem{ajay2018pervasive}
J.~Ajay, C.~Song, A.~Wang, J.~Langan, Z.~Li, and W.~Xu, ``A pervasive and
  sensor-free deep learning system for parkinsonian gait analysis,'' in {\em
  2018 IEEE EMBS International Conference on Biomedical \& Health Informatics
  (BHI)}, pp.~108--111, IEEE, 2018.

\bibitem{el2020deep}
I.~El~Maachi, G.-A. Bilodeau, and W.~Bouachir, ``Deep 1d-convnet for accurate
  parkinson disease detection and severity prediction from gait,'' {\em Expert
  Systems with Applications}, vol.~143, p.~113075, 2020.

\bibitem{zou2020deep}
Q.~Zou, Y.~Wang, Q.~Wang, Y.~Zhao, and Q.~Li, ``Deep learning-based gait
  recognition using smartphones in the wild,'' {\em IEEE Transactions on
  Information Forensics and Security}, vol.~15, pp.~3197--3212, 2020.

\bibitem{zhao2021multimodal}
A.~Zhao, J.~Li, J.~Dong, L.~Qi, Q.~Zhang, N.~Li, X.~Wang, and H.~Zhou,
  ``Multimodal gait recognition for neurodegenerative diseases,'' {\em IEEE
  transactions on cybernetics}, vol.~52, no.~9, pp.~9439--9453, 2021.

\bibitem{balaji2021automatic}
E.~Balaji, D.~Brindha, V.~K. Elumalai, and R.~Vikrama, ``Automatic and
  non-invasive parkinson’s disease diagnosis and severity rating using lstm
  network,'' {\em Applied Soft Computing}, vol.~108, p.~107463, 2021.

\bibitem{kumar2023parkinson}
K.~V. S.~S. Kumar, I.~Sirisha, K.~Vathsalya, and K.~K. V.~V. Vamsi, ``Parkinson
  disease diagnosis and severity rating prediction based on gait analysis using
  deep learning,'' {\em International Research Journal on Advanced Science
  Hub}, vol.~5, no.~05, pp.~418--425, 2023.

\bibitem{info14020119}
M.~Chen, T.~Ren, P.~Sun, J.~Wu, J.~Zhang, and A.~Zhao, ``Fuselgnet: Fusion of
  local and global information for detection of parkinson's disease,'' {\em
  Information}, vol.~14, no.~2, 2023.

\bibitem{9956330}
D.~M.~D. Nguyen, M.~Miah, G.-A. Bilodeau, and W.~Bouachir, ``Transformers for
  1d signals in parkinson’s disease detection from gait,'' in {\em 2022 26th
  International Conference on Pattern Recognition (ICPR)}, pp.~5089--5095,
  2022.

\bibitem{CHERIET2023107344}
M.~Cheriet, V.~Dentamaro, M.~Hamdan, D.~Impedovo, and G.~Pirlo, ``Multi-speed
  transformer network for neurodegenerative disease assessment and activity
  recognition,'' {\em Computer Methods and Programs in Biomedicine}, vol.~230,
  p.~107344, 2023.

\bibitem{sun2022transformer}
H.-J. Sun and Z.-G. Zhang, ``Transformer-based severity detection of
  parkinson's symptoms from gait,'' in {\em 2022 15th International Congress on
  Image and Signal Processing, BioMedical Engineering and Informatics
  (CISP-BMEI)}, pp.~1--5, IEEE, 2022.

\bibitem{naimi2023hct}
S.~Naimi, W.~Bouachir, and G.-A. Bilodeau, ``Hct: Hybrid convnet-transformer
  for parkinson's disease detection and severity prediction from gait,'' in
  {\em 2023 International Conference on Machine Learning and Applications
  (ICMLA)}, pp.~814--819, IEEE, 2023.

\bibitem{wu2022multi}
H.~Wu and A.~Zhao, ``Multi-level fine-tuned transformer for gait recognition,''
  in {\em 2022 International Conference on Virtual Reality, Human-Computer
  Interaction and Artificial Intelligence (VRHCIAI)}, pp.~83--89, IEEE, 2022.

\bibitem{wu2023attention}
H.~Wu, Y.~Liu, H.~Yang, Z.~Xie, X.~Chen, M.~Wen, and A.~Zhao, ``An
  attention-based temporal network for parkinson's disease severity rating
  using gait signals,'' {\em KSII Transactions on Internet and Information
  Systems (TIIS)}, vol.~17, no.~10, pp.~2627--2642, 2023.

\bibitem{2022Sensor}
N.~Kour, S.~Gupta, and S.~Arora, ``Sensor technology with gait as a diagnostic
  tool for assessment of parkinson's disease: a survey,'' {\em Multimedia Tools
  and Applications}, pp.~1--37, 2022.

\bibitem{naimi20241d}
S.~Naimi, W.~Bouachir, and G.-A. Bilodeau, ``1d-convolutional transformer for
  parkinson disease diagnosis from gait,'' {\em Neural Computing and
  Applications}, vol.~36, no.~4, pp.~1947--1957, 2024.

\bibitem{chen2023fuselgnet}
M.~Chen, T.~Ren, P.~Sun, J.~Wu, J.~Zhang, and A.~Zhao, ``Fuselgnet: Fusion of
  local and global information for detection of parkinson’s disease,'' {\em
  Information}, vol.~14, no.~2, p.~119, 2023.

\bibitem{9087433}
A.~Moshkova, A.~Samorodov, N.~Voinova, A.~Volkov, E.~Ivanova, and E.~Fedotova,
  ``Parkinson’s disease detection by using machine learning algorithms and
  hand movement signal from leapmotion sensor,'' in {\em 2020 26th Conference
  of Open Innovations Association (FRUCT)}, vol.~1, pp.~321--327, 2020.

\bibitem{9411721}
L.~Tong, J.~He, and L.~Peng, ``Cnn-based pd hand tremor detection using
  inertial sensors,'' {\em IEEE Sensors Letters}, vol.~5, no.~7, pp.~1--4,
  2021.

\bibitem{2022Two}
A.~Zhao and J.~Li, ``Two-channel lstm for severity rating of parkinson's
  disease using 3d trajectory of hand motion,'' {\em Multimedia Tools and
  Applications}, vol.~81, no.~23, pp.~33851--33866, 2022.

\bibitem{ma2021retracted}
B.~Ma, F.~Zhang, and B.~Ma, ``[retracted] self-attention-guided recurrent
  neural network and motion perception for intelligent prediction of chronic
  diseases,'' {\em Journal of Healthcare Engineering}, vol.~2021, no.~1,
  p.~6382619, 2021.

\bibitem{guo2022tree}
R.~Guo, H.~Li, C.~Zhang, and X.~Qian, ``A tree-structure-guided graph
  convolutional network with contrastive learning for the assessment of
  parkinsonian hand movements,'' {\em Medical Image Analysis}, vol.~81,
  p.~102560, 2022.

\bibitem{peng2024multi}
X.~Peng, Y.~Zhao, Z.~Li, X.~Wang, F.~Nan, Z.~Zhao, Y.~Yang, and P.~Yang,
  ``Multi-scale and multi-level feature assessment framework for classification
  of parkinson’s disease state from short-term motor tasks,'' {\em IEEE
  Transactions on Biomedical Engineering}, 2024.

\bibitem{zhao2024selecting}
Y.~Zhao, X.~Wang, X.~Peng, Z.~Li, F.~Nan, M.~Zhou, J.~Qi, Y.~Yang, Z.~Zhao,
  L.~Xu, {\em et~al.}, ``Selecting and evaluating key mds-updrs activities
  using wearable devices for parkinson's disease self-assessment,'' {\em IEEE
  Journal of Selected Areas in Sensors}, 2024.

\bibitem{gazda2021multiple}
M.~Gazda, M.~Hire{\v{s}}, and P.~Drot{\'a}r, ``Multiple-fine-tuned
  convolutional neural networks for parkinson’s disease diagnosis from
  offline handwriting,'' {\em IEEE Transactions on Systems, Man, and
  Cybernetics: Systems}, vol.~52, no.~1, pp.~78--89, 2021.

\bibitem{wang2023coordinate}
N.~Wang, X.~Niu, Y.~Yuan, Y.~Sun, R.~Li, G.~You, and A.~Zhao, ``A coordinate
  attention enhanced swin transformer for handwriting recognition of
  parkinson's disease,'' {\em IET Image Processing}, vol.~17, no.~9,
  pp.~2686--2697, 2023.

\bibitem{ma2022feature}
C.~Ma, P.~Zhang, L.~Pan, X.~Li, C.~Yin, A.~Li, R.~Zong, and Z.~Zhang, ``A
  feature fusion sequence learning approach for quantitative analysis of tremor
  symptoms based on digital handwriting,'' {\em Expert Systems with
  Applications}, vol.~203, p.~117400, 2022.

\bibitem{sabapathy2022competent}
S.~Sabapathy, S.~Maruthu, S.~K. Krishnadhas, A.~K. Tamilarasan, and
  N.~Raghavan, ``Competent and affordable rehabilitation robots for nervous
  system disorders powered with dynamic cnn and hmm,'' {\em Intelligent Systems
  for Rehabilitation Engineering}, pp.~57--93, 2022.

\bibitem{SADHU2022100351}
S.~Sadhu, D.~Solanki, N.~Constant, V.~Ravichandran, G.~Cay, M.~J. Saikia,
  U.~Akbar, and K.~Mankodiya, ``Towards a telehealth infrastructure supported
  by machine learning on edge/fog for parkinson's movement screening,'' {\em
  Smart Health}, vol.~26, p.~100351, 2022.

\bibitem{electronics8080907}
M.~Gil-Martín, J.~M. Montero, and R.~San-Segundo, ``Parkinson’s disease
  detection from drawing movements using convolutional neural networks,'' {\em
  Electronics}, vol.~8, no.~8, 2019.

\bibitem{9359526}
A.~Zhao, J.~Dong, J.~Li, L.~Qi, and H.~Zhou, ``Associated spatio-temporal
  capsule network for gait recognition,'' {\em IEEE Transactions on
  Multimedia}, vol.~24, pp.~846--860, 2022.

\bibitem{9719945}
M.~Mohaghegh and J.~Gascon, ``Identifying parkinson’s disease using
  multimodal approach and deep learning,'' in {\em 2021 6th International
  Conference on Innovative Technology in Intelligent System and Industrial
  Applications (CITISIA)}, pp.~1--6, 2021.

\bibitem{MA2021102849}
J.~Ma, Y.~Zhang, Y.~Li, L.~Zhou, L.~Qin, Y.~Zeng, P.~Wang, and Y.~Lei, ``Deep
  dual-side learning ensemble model for parkinson speech recognition,'' {\em
  Biomedical Signal Processing and Control}, vol.~69, p.~102849, 2021.

\bibitem{Zhao2023}
A.~Zhao, Y.~Wang, and L.~Jianbo, ``Transferable self-supervised instance
  learning for sleep recognition,'' {\em IEEE Transactions on Multimedia},
  vol.~PP, pp.~1--1, 05 2023.

\bibitem{wang2021hierarchical}
C.~Wang, L.~Peng, Z.-G. Hou, Y.~Li, Y.~Tan, and H.~Hao, ``A hierarchical
  architecture for multisymptom assessment of early parkinson’s disease via
  wearable sensors,'' {\em IEEE Transactions on Cognitive and Developmental
  Systems}, vol.~14, no.~4, pp.~1553--1563, 2021.

\bibitem{junaid2023explainable}
M.~Junaid, S.~Ali, F.~Eid, S.~El-Sappagh, and T.~Abuhmed, ``Explainable machine
  learning models based on multimodal time-series data for the early detection
  of parkinson’s disease,'' {\em Computer Methods and Programs in
  Biomedicine}, vol.~234, p.~107495, 2023.

\bibitem{pahuja2022deep}
G.~Pahuja and B.~Prasad, ``Deep learning architectures for parkinson's disease
  detection by using multi-modal features,'' {\em Computers in Biology and
  Medicine}, vol.~146, p.~105610, 2022.

\bibitem{li2023multimodal}
J.~Li, W.~Liang, X.~Yin, J.~Li, and W.~Guan, ``Multimodal gait abnormality
  recognition using a convolutional neural network--bidirectional long
  short-term memory (cnn-bilstm) network based on multi-sensor data fusion,''
  {\em Sensors}, vol.~23, no.~22, p.~9101, 2023.

\bibitem{faiem2024assessment}
N.~Faiem, T.~Asuroglu, K.~Acici, A.~Kallonen, and M.~Van~Gils, ``Assessment of
  parkinson’s disease severity using gait data: A deep learning-based
  multimodal approach,'' in {\em Nordic Conference on Digital Health and
  Wireless Solutions}, pp.~29--48, Springer, 2024.

\bibitem{wang2024prediction}
B.~Wang, X.~Hu, R.~Ge, C.~Xu, J.~Zhang, Z.~Gao, S.~Zhao, and K.~Polat,
  ``Prediction of freezing of gait in parkinson’s disease based on
  multi-channel time-series neural network,'' {\em Artificial Intelligence in
  Medicine}, vol.~154, p.~102932, 2024.

\bibitem{zhao2023spatio}
A.~Zhao, H.~Wu, M.~Chen, and N.~Wang, ``A spatio-temporal siamese neural
  network for multimodal handwriting abnormality screening of parkinson’s
  disease,'' {\em International Journal of Intelligent Systems}, vol.~2023,
  no.~1, p.~9921809, 2023.

\bibitem{xue2024ai}
C.~Xue, S.~S. Kowshik, D.~Lteif, S.~Puducheri, V.~H. Jasodanand, O.~T. Zhou,
  A.~S. Walia, O.~B. Guney, J.~D. Zhang, S.~T. Pham, {\em et~al.}, ``Ai-based
  differential diagnosis of dementia etiologies on multimodal data,'' {\em
  Nature Medicine}, pp.~1--13, 2024.

\bibitem{misc_parkinsons_174}
M.~Little, ``{Parkinsons}.'' UCI Machine Learning Repository, 2008.
\newblock {DOI}: https://doi.org/10.24432/C59C74.

\bibitem{misc_park301}
K.~Olcay, S.~Betul, I.~M., S.~C., S.~Ahmet, and G.~Fikret, ``{Parkinson's
  Speech with Multiple Types of Sound Recordings}.'' UCI Machine Learning
  Repository, 2014.
\newblock {DOI}: https://doi.org/10.24432/C5NC8M.

\bibitem{hagen_jaeger_2019_2867216}
H.~Jaeger, D.~Trivedi, and M.~Stadtschnitzer, ``{Mobile Device Voice Recordings
  at King's College London (MDVR-KCL) from both early and advanced Parkinson's
  disease patients and healthy controls},'' May 2019.

\bibitem{aw6b-tg17-19}
G.~Dimauro and F.~Girardi, ``Italian parkinson's voice and speech,'' 2019.

\bibitem{misc470}
S.~C., S.~Gorkem, G.~Aysegul, N.~Hatice, and S.~Betul, ``{Parkinson's Disease
  Classification}.'' UCI Machine Learning Repository, 2018.
\newblock {DOI}: https://doi.org/10.24432/C5MS4X.

\bibitem{7813053}
C.~R. {Pereira}, S.~A.~T. {Weber}, C.~{Hook}, G.~H. {Rosa}, and J.~P. {Papa},
  ``Deep learning-aided parkinson's disease diagnosis from handwritten
  dynamics,'' in {\em 2016 29th SIBGRAPI Conference on Graphics, Patterns and
  Images (SIBGRAPI)}, pp.~340--346, 2016.

\bibitem{2014Improved}
M.~E. Isenkul, B.~E. Sakar, and O.~Kursun, ``Improved spiral test using
  digitized graphics tablet for monitoring parkinson's disease,'' in {\em The
  2nd International Conference on E-Health and TeleMedicine-ICEHTM 2014},
  pp.~1--1, 2014.

\bibitem{misc_58}
K.~Olcay, ``{Improved Spiral Test Using Digitized Graphics Tablet for
  Monitoring Parkinson's Disease}.'' UCI Machine Learning Repository, 2016.
\newblock {DOI}: https://doi.org/10.24432/C5HW3N.

\bibitem{10.1371/journal.pone.0188226}
W.~R. Adams, ``High-accuracy detection of early parkinson's disease using
  multiple characteristics of finger movement while typing,'' {\em PLOS ONE},
  vol.~12, pp.~1--20, 11 2017.

\bibitem{10.1038/srep34468}
L.~Giancardo, ``Computer keyboard interaction as an indicator of early
  parkinson's disease,'' {\em Scientific reports}, vol.~6, p.~34468, 10 2016.

\bibitem{s41531-023-00625-7}
V.~J., B.~A., F.~M., van Alen, C.M., L.~Plagwitz, and W.~T., ``Machine learning
  in the parkinson's disease smartwatch (pads) dataset.,'' {\em npj
  Parkinsons}, vol.~9, p.~6, 1 2024.

\bibitem{physionet-PDgait}
NIGMS, ``physionet pdgait database.''
  \url{https://physionet.org/content/gaitpdb/1.0.0/}.

\bibitem{physionet-ndd}
NIGMS, ``physionet ndds database.''
  \url{https://physionet.org/physiobank/database/gaitndd/}.

\bibitem{Salomon2024article}
A.~Salomon, E.~Gazit, P.~Ginis, B.~Urazalinov, H.~Takoi, T.~Yamaguchi, S.~Goda,
  D.~Lander, J.~Lacombe, A.~Sinha, A.~Nieuwboer, L.~Kirsch, R.~Holbrook,
  B.~Manor, and J.~Hausdorff, ``A machine learning contest enhances automated
  freezing of gait detection and reveals time-of-day effects,'' {\em Nature
  Communications}, vol.~15, 06 2024.

\bibitem{misc_daphnet_245}
R.~Daniel, P.~Meir, and H.~Jeff, ``{Daphnet Freezing of Gait}.'' UCI Machine
  Learning Repository, 2013.
\newblock {DOI}: https://doi.org/10.24432/C56K78.

\bibitem{physionet-multimodal}
NIGMS, ``physionet multimodal database.''
  \url{https://physionet.org/content/multi-gait-posture/1.0.0/}.

\bibitem{morgan2023multimodal}
C.~Morgan, E.~L. Tonkin, A.~Masullo, F.~Jovan, A.~Sikdar, P.~Khaire,
  M.~Mirmehdi, R.~McConville, G.~J. Tourte, A.~Whone, {\em et~al.}, ``A
  multimodal dataset of real world mobility activities in parkinson’s
  disease,'' {\em Scientific data}, vol.~10, no.~1, p.~918, 2023.

\bibitem{li2021multimodal}
H.~Li, ``Multimodal dataset of freezing of gait in parkinson’s disease,''
  {\em Mendeley Data}, vol.~3, p.~2021, 2021.

\bibitem{ribeiro2022public}
C.~Ribeiro De~Souza, R.~Miao, J.~{\'A}vila De~Oliveira, A.~Cristina De
  Lima-Pardini, D.~Fragoso De~Campos, C.~Silva-Batista, L.~Teixeira, S.~Shokur,
  B.~Mohamed, and D.~B. Coelho, ``A public data set of videos, inertial
  measurement unit, and clinical scales of freezing of gait in individuals with
  parkinson's disease during a turning-in-place task,'' {\em Frontiers in
  Neuroscience}, vol.~16, p.~832463, 2022.

\end{thebibliography}
\end{document}